\documentclass{article}
\usepackage{graphicx} % Required for inserting images

\usepackage{times}  % DO NOT CHANGE THIS
\usepackage{helvet}  % DO NOT CHANGE THIS
\usepackage{courier}  % DO NOT CHANGE THIS
\usepackage[hyphens]{url}  % DO NOT CHANGE THIS
\usepackage{graphicx} % DO NOT CHANGE THIS
\usepackage{caption} % DO NOT CHANGE THIS AND DO NOT ADD ANY OPTIONS TO IT
\usepackage{algorithm}
\usepackage{algorithmic}

\usepackage{amsmath}
\usepackage{amsfonts}
\usepackage{amssymb}
\usepackage{multirow}
\usepackage{booktabs}
\usepackage{tabularx}
\usepackage{hyperref} % aaai not allowed to use this
\usepackage{longtable}
\usepackage{makecell}
\usepackage{url}
\usepackage{float}

\def \eval#1{\textit{#1}}
\def \best#1{\textbf{#1}}
\def \second#1{\underline{#1}}
\hypersetup{colorlinks=true}
\newcommand{\source}[1]{\textsc{\footnotesize #1}}

\usepackage{newfloat}
\usepackage{listings}

\title{\textbf{LD-DETR: Loop Decoder DEtection TRansformer for Video Moment Retrieval and Highlight Detection}}
\author{\textbf{Pengcheng Zhao,}
    \textbf{Zhixian He,}\\
    \textbf{Fuwei Zhang,} 
    \textbf{Shujin Lin\thanks{Corresponding author.},} 
    \textbf{Fan Zhou}\\
    Sun Yat-Sen University, Guangzhou, China\\
    qingchen239@gmail.com, hezhx29@mail2.sysu.edu.cn, \\
    zhangfw5@mail2.sysu.edu.cn, linshjin@mail.sysu.edu.cn, \\
    isszf@mail.sysu.edu.cn
}
\date{}

\begin{document}

\maketitle

\begin{abstract}
Video Moment Retrieval and Highlight Detection aim to find corresponding content in the video based on a text query. Existing models usually first use contrastive learning methods to align video and text features, then fuse and extract multimodal information, and finally use a Transformer Decoder to decode multimodal information. However, existing methods face several issues: 
% \begin{enumerate}
%     \item Overlapping semantic information between different samples in the dataset hinders the model's multimodal aligning performance;
%     \item Existing models are not able to efficiently extract local features of the video;
%     \item The Transformer Decoder used by the existing model cannot adequately decode multimodal features. To address the above issues, we proposed the LD-DETR model for Video Moment Retrieval and Highlight Detection tasks.
% \end{enumerate}
(1) Overlapping semantic information between different samples in the dataset hinders the model's multimodal aligning performance; (2) Existing models are not able to efficiently extract local features of the video; (3) The Transformer Decoder used by the existing model cannot adequately decode multimodal features. To address the above issues, we proposed the LD-DETR model for Video Moment Retrieval and Highlight Detection tasks. 
Specifically,  
% \begin{itemize}
%     \item we first distilled the similarity matrix into the identity matrix to mitigate the impact of overlapping semantic information.
%     \item Then, we designed a method that enables convolutional layers to extract multimodal local features more efficiently.
%     \item Finally, we fed the output of the Transformer Decoder back into itself to adequately decode multimodal information. 
% \end{itemize}
we first distilled the similarity matrix into the identity matrix to mitigate the impact of overlapping semantic information. Then, we designed a method that enables convolutional layers to extract multimodal local features more efficiently. Finally, we fed the output of the Transformer Decoder back into itself to adequately decode multimodal information. 
We evaluated LD-DETR on four public benchmarks and conducted extensive experiments to demonstrate the superiority and effectiveness of our approach. Our model outperforms the State-Of-The-Art models on QVHighlight, Charades-STA and TACoS datasets. Our code is available at \url{https://github.com/qingchen239/ld-detr}. 
\end{abstract}

\clearpage

\section{Introduction}

Video Moment Retrieval aims to identify specific moments within a video that correspond to a given text query~\cite{liu2015multi, anne2017localizing, escorcia2019temporal, gao2017tall, liu2018cross, shimomoto2022towards, zeng2023temporally, zhang2019man}. 
% liu2015multi, anne2017localizing, escorcia2019temporal, gao2017tall, liu2018cross, shimomoto2022towards, zeng2023temporally, zhang2019man
Highlight Detection evaluates the degree of relevance of different time clips to the text~\cite{yao2016highlight, zhang2016video, badamdorj2022contrastive, mahasseni2017unsupervised, wei2022learning, gygli2016video2gif, xiong2019less, yu2018deep}. 
% yao2016highlight, zhang2016video, badamdorj2022contrastive, mahasseni2017unsupervised, wei2022learning, gygli2016video2gif, xiong2019less, yu2018deep
With the development of digital devices and platforms, users' demand for video content has increased significantly. Finding interesting clips in videos quickly and accurately has become an important requirement. Therefore, the research on Video Moment Retrieval and Highlight Detection has attracted widespread attention. 

Existing models on Video Moment Retrieval and Highlight Detection often use contrastive learning to align video and text features~\cite{moon2023query, sun2024tr, moon2023correlation, liu2024r, xiao2024bridging}, 
% For instance, QD-DETR~\cite{moon2023query} employs contrastive learning by displacing video-text pairs to create negative samples. TR-DETR~\cite{sun2024tr} and CG-DETR~\cite{moon2023correlation} extract global features for each video and text, aligning them by ensuring their correlation matrices approximate the identity matrix. R\textsuperscript{2}-Tuning~\cite{liu2024r} uses multiple contrastive learning layer for this purpose. Existing models usually 
use attention mechanism to fuse and extract multimodal information~\cite{liu2022umt, moon2023query, sun2024tr, lei2021detecting, li2024momentdiff, moon2023correlation}, 
% UMT~\cite{liu2022umt} and QD-DETR~\cite{moon2023query} introduced a Cross-Attentive Transformer Encoder to effectively fuse multimodal features. CG-DETR~\cite{moon2023correlation} advanced this approach by proposing an Adaptive Cross-Attention mechanism with dummy tokens, which enhances the flexibility and effectiveness of feature fusion. TR-DETR~\cite{sun2024tr} introduced Visual Feature Refinement to filter out irrelevant video information, ensuring that the fused features are more relevant to the text. Most existing models 
and use a Transformer Decoder with a zero matrix as $query$~\cite{zheng2020end, lei2021detecting} to decode the fused multimodal information~\cite{liu2022umt, moon2023query, sun2024tr, lei2021detecting, li2024momentdiff, 1024497142.nh, moon2023correlation}. 
% BM-DETR~\cite{jung2023overcoming} introduces a learnable $query$ for the decoder. MomentDiff~\cite{li2024momentdiff} uses a diffusion model decoder to improve the ability to express. UVCOM~\cite{xiao2024bridging} uses the cross-attention of text and video features as the $query$ of the Transformer Decoder. 

However, the existing methods face several issues:
\begin{enumerate}
    % (1) Increasing the number of contrastive learning samples, for example by increasing the batch size, can improve model performance ~\cite{he2020momentum, caron2020unsupervised, chen2020simple, grill2020bootstrap}. However, when GPU resources are limited, simply expanding the batch size is not feasible, because it occupies a large amount of GPU memory. 
    % (1) Same semantic information inevitably exists in different training samples~\cite{jung2023overcoming}. This overlapping semantic information makes the negative samples in contrastive learning correlate to a certain degree rather than completely irrelevant. However, existing models treat them equally as negative samples, which reduces the model's performance. 
    \item In contrastive learning, methods generally regard features from different samples as negative samples~\cite{sun2024tr, moon2023correlation}, but the same semantic information inevitably appears in different samples (\textit{e.g.}, “man eats” and “man drinks” both have the same information “man”)~\cite{jung2023overcoming, liu2024towards}. Treating them simply as completely negative samples hinders the performance of multimodal alignment. 
    \item The corresponding contents are usually small parts of the video and have strong local correlations, but the current models 
    % focus more on removing features in video features not related to text features, and ignore the temporal structure of the video itself. They extract video features with a Transformer Encoder~\cite{vaswani2017attention}, whose global attention calculation~\cite{bahdanau2014neural} dilutes the weight of local details, causing the model to 
    ignore extracting local features of the video. An intuitive approach is to use convolutional layers~\cite{lecun1998gradient, krizhevsky2012imagenet} to extract local features~\cite{xiao2024bridging, 1024497142.nh}, but simply stacking convolutional layers does not improve the performance of existing models.
    \item Researches~\cite{yang2024task, liu2022dab, zhu2020deformable, gao2021fast, meng2021conditional, wang2022anchor, yao2021efficient} have shown that the Transformer Decoder inadequately processes the retrieval. A bigger decoder might improve the model's retrieval abilities but also risks overfitting. 
\end{enumerate}

\begin{figure}[!h]
\centering
\includegraphics[width=0.8\textwidth]{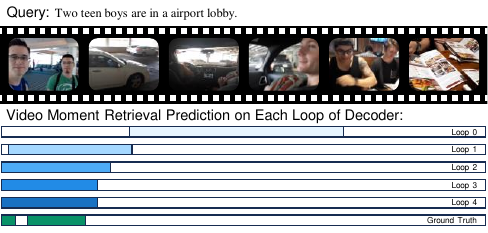}
\caption{\textbf{Loop Decoder makes Video Moment Retrieval more accurate. }We visualize the Video Moment Retrieval results corresponding to the output of Loop Decoder at each loop. As the number of loops increases, the prediction gets closer and closer to the ground truth. This experiment uses Moment-DETR~\cite{lei2021detecting} as the baseline. }
\label{fig:loop_decoder_illustration}
\end{figure}

To address these issues, we proposed the Loop Decoder DEtection TRansformer model (LD-DETR): 
\begin{enumerate}
    % (1) Inspired by MoCo~\cite{he2020momentum}, we update the encoder with momentum to keep the consistency of the extracted global features. We propose the \textit{Distill Align} to store features in a queue to increase the number of features involved in contrastive learning, 
    % without excessive GPU memory usage
    \item Different from other methods that used an identity matrix as the target of contrastive learning, we distilled a matrix representing the correlation between samples into the identity matrix to mitigate the impact of overlapping semantic information. 
    \item Convolutional layers have small receptive fields. This feature allows the network to capture local information. We designed a method that enables stacked convolutional layers to extract multimodal local features more efficiently. 
    \item Research shows that when the Transformer Decoder’s $query$ carries target information, the decoder can better decode the input information~\cite{liu2022dab}, and the output of the Transformer Decoder carries exactly the target information. Inspired by this, we fed the Transformer Decoder's output back into itself as the $query$, as shown in Figure \ref{fig:loop_decoder_illustration}, to enhance its ability to decode multimodal fusion information adequately without increasing the risk of overfitting. 
\end{enumerate}

We evaluated LD-DETR on four public benchmarks and conducted extensive experiments to demonstrate the superiority and effectiveness of our approach. 

The main contributions of our work are summarized below:
\begin{itemize}
    \item We introduced a plug-and-play method, \textit{Distill Align}, which
    % can involve more features in contrastive learning without occupying too much GPU memory and 
    takes the impact of overlapping semantic information between training samples into account when aligning multimodal features to improve the performance of the model.
    \item We introduced \textit{Convolutional Fuser} to better extract local features in videos and achieved excellent results. 
    % We are the first to incorporate existing image recognition networks as methods into the model to improve the performance of Video Moment Retrieval and Highlight Detection tasks.
    \item We proposed a plug-and-play method, \textit{Loop Decoder}, which improves the decoder's ability to decode multimodal fusion information adequately without causing overfitting.
    \item Based on the above methods, we designed a model LD-DETR for Video Moment Retrieval and Highlight Detection tasks, and verified its advancedness and effectiveness on multiple datasets. Our model outperforms the state-of-the-art models on QVHighlight, Charades-STA and TACoS datasets. 
\end{itemize}

\clearpage

\section{Related Work}

\begin{figure*}[!h]
\centering
\includegraphics[width=1\textwidth]{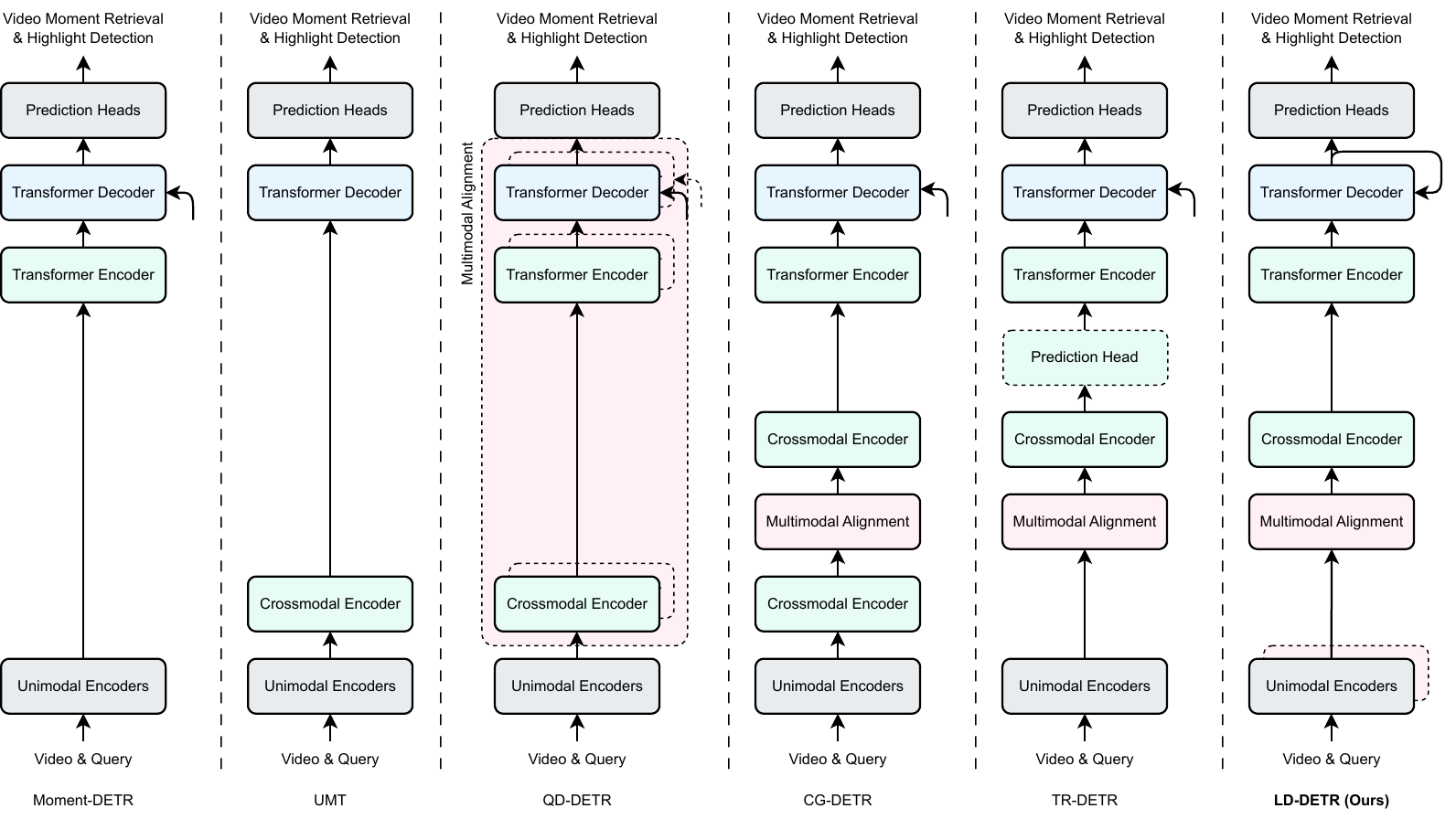}
\caption{\textbf{The overall framework of several recent proposed models. }Those methods mainly improve the model from three perspectives: Aligning Multimodal Features, Fusing and Extracting Multimodal Features, and Decoding Multimodal Information. 
}
\label{fig:related_work}
\end{figure*}

Since the QVHighlight dataset~\cite{lei2021detecting} was proposed, Video Moment Retrieval and Highlight Detection tasks have been jointly studied, and many models based on DEtection TRansformer (DETR)~\cite{zheng2020end} have been proposed~\cite{lei2021detecting, liu2022umt, jung2023overcoming, moon2023query, moon2023correlation, lee2023bam, sun2024tr, li2024momentdiff, xiao2024bridging, yang2024task}. As shown in Figure \ref{fig:related_work}, these methods mainly improve the model from three perspectives: Aligning Multimodal Features, Fusing and Extracting Multimodal Features, and Decoding Multimodal Information. 

% Video Moment Retrieval aims to identify specific moments within a video that correspond to a given text query. Highlight Detection, on the other hand, evaluates the degree of relevance of different time clips to the text. Although these two works have a lot in common, they have been studied separately for a long time. Until recently, Lei, Berg, and Bansal realized the commonality of these two works and proposed QVHighlight~\cite{lei2021detecting}, currently the only dataset that can applied to both tasks, and the first model Moment-DETR~\cite{lei2021detecting} that is applied to both tasks. 
% % Moment-DETR is based on DETR~\cite{zheng2020end}. 
% Subsequently, several DETR-based~\cite{zheng2020end} Video Moment Retrieval and Highlight Detection models were proposed, and gradually became the mainstream of these two tasks. 

\subsection{Contrastive Learning and Aligning Multimodal Features}

% Contrastive learning is a machine learning method that trains a model by comparing and contrasting different samples to improve its performance in distinguishing features and recognizing patterns. InstDisc~\cite{wu2018unsupervised} introduces a memory bank mechanism to store the feature vectors of all instances for contrastive learning, and updates them in real time as training progresses. CMC~\cite{tian2020contrastive} uses contrastive learning to map different perspectives of the same image into similar semantic spaces, demonstrating the feasibility of contrastive learning to align multimodal information. 

Contrastive learning is a machine learning method that compares and contrasts different samples to improve its performance in distinguishing features and recognizing patterns. CMC~\cite{tian2020contrastive} uses contrastive learning to map different perspectives of the same image into similar semantic spaces, demonstrating the feasibility of aligning multimodal information. MoCo~\cite{he2020momentum} obtains the features of negative samples through a momentum-updated encoder and stores them in a queue to increase the number of samples involved in contrastive learning. SimCLR~\cite{chen2020simple} introduces a learnable nonlinear transformation between the representation and the contrastive loss substantially improves the quality of the learned representations. SimSiam~\cite{chen2021exploring} directly maximizes the similarity of one image’s two views, using neither negative pairs nor a momentum encoder. ALBEF~\cite{li2021align} learns from pseudo-targets produced by a momentum model to improve learning from noisy web data. CLIP~\cite{radford2021learning} extracts global features for each image and text, aligning them by ensuring their correlation matrices approximate the identity matrix. 

Existing Video Moment Retrieval and Highlight Detection models generally use CLIP-like methods to align video and text features. TR-DETR~\cite{sun2024tr} takes the average of the video and text features encoded by the unimodal encoder as the global feature of the sample, then aligns the multimodal features using a CLIP-like method before mixing the multimodal features. CG-DETR~\cite{moon2023correlation} uses two different encoders to extract global features of videos and texts of positive and negative samples respectively, and aligns them separately through a CLIP-like method. It also distills a similarity map to adjust the magnitude of the attention map in the cross-attention layer. In these CLIP-like methods, the number of features involved in contrastive learning is limited by the batch size. 

BM-DETR~\cite{jung2023overcoming} proposes a weak alignment problem, the overlapping semantic information between different samples in the dataset reduces the model performance, and solves it at the multimodal feature fusion level. 
% We believe that this problem may also reduce multimodal alignment, and we will solve it at the multimodal alignment level. 

\subsection{Fusing and Extracting Multimodal Features}

Moment-DETR~\cite{lei2021detecting} simply concats the video and text features and feeds them into the Transformer Encoder. UMT~\cite{liu2022umt} and QD-DETR~\cite{moon2023query} propose a method in which Cross-Attentive Transformer Encoder uses text features to encode video features in order to remove information in video features that is irrelevant to the query text. CG-DETR~\cite{moon2023correlation} goes a step further by concatenating noise onto text features based on the Cross-Attentive Transformer Encoder to better remove irrelevant information. TR-DETR~\cite{sun2024tr} introduces Visual Feature Refinement to filter out irrelevant video information. These existing methods~\cite{liu2022umt, moon2023query, sun2024tr, lei2021detecting, li2024momentdiff, 1024497142.nh, moon2023correlation} only focus on removing features that are not related to text in video features, while ignoring the temporal structure of the video itself, and leave the task of extracting video features to a Transformer Encoder~\cite{vaswani2017attention}. But the Transformer Encoder's global attention calculation~\cite{bahdanau2014neural} dilutes the weight of local details, causes the model to ignore the extraction of local features of the video. UVCOM~\cite{xiao2024bridging} proposes the Comprehensive Integration Module method, which has a convolutional layer to extract local multimodal features. CDIM~\cite{1024497142.nh} also proposes a Cross-Modal Convolutional Interaction method which stacks dilated convolutional layers to enhance the model's perceptive capability. 

\subsection{Decoding Multimodal Information}

Most existing models are based on a model called DETR~\cite{zheng2020end} that is applied to object detection. Researches~\cite{yang2024task, liu2022dab, zhu2020deformable, gao2021fast, meng2021conditional, wang2022anchor, yao2021efficient} have shown that Transformer Decoders inadequately process fused multimodal information, and propose solutions suitable for object detection tasks. MomentDiff~\cite{li2024momentdiff} notices that unevenly distributed training samples lead to a lack of generalization of the model, and introduces a diffusion model method into the decoder to solve this problem. UVCOM~\cite{xiao2024bridging} uses the cross-attention of text and video features as the $query$ of the Transformer Decoder. TaskWeave~\cite{yang2024task} uses a decoder following DAB-DETR~\cite{liu2022dab} to decode Video Moment Retrieval features, and uses another decoder to decode Highlight Detection features. 

\clearpage

\section{Model and Methods}

\subsection{Overview of LD-DETR}

Given a video containing $t$ clips, and a text query containing $n$ tokens, the goal of Video Moment Retrieval and Highlight Detection is to find all moments $\{ c_{\textnormal{m}}, w_{\textnormal{m}} \}_{\textnormal{m}}^m$ related to the text query in the video, where $c_{\textnormal{m}}$ and $w_{\textnormal{m}}$ represent the center time and duration length of the $\textnormal{m}$-th moment and $m$ is the total number of predicted moments, and predict the saliency scores $s\in\mathbb{R}^t$ of all moments at the clip-level. 

\begin{figure*}[!h]
\centering
\includegraphics[width=1\textwidth]{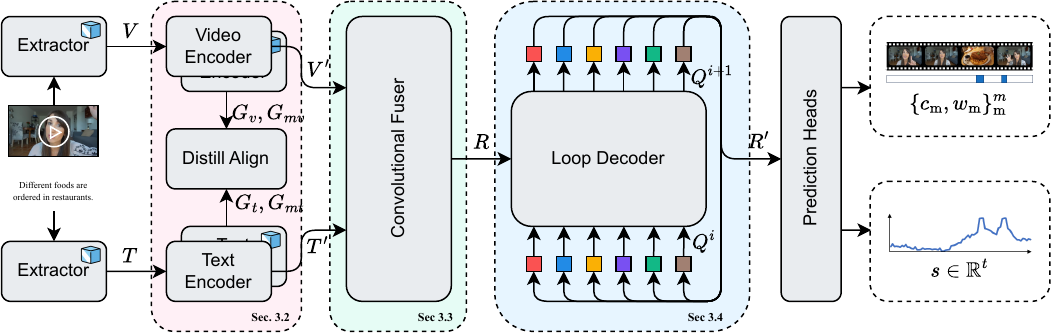}
\caption{\textbf{The overall framework of our model LD-DETR. }
% We use two pre-trained extractors to extract features from video and text respectively. We use two unimodal encoders to map the extracted video and text features to the same latent space. Then we additionally add two momentum unimodal encoders and Distill Align method to ensure that the features are mapped to the same space. Then we feed the video and text features into the multimodal fusion and extraction method. We feed the fused and extracted multimodal features into the Loop Decoder to obtain the decoded features. Finally, we feed the decoded features into the prediction heads to obtain the video moment retrieval and highlight detection. 
For the methods marked with an ice cube, their parameters are not affected by gradient descent during training. 
}
\label{fig:model}
\end{figure*}

% The overall framework of our model is derived from the Transformer Encoder-decoder architecture. 
As shown in Figure \ref{fig:model}, the LD-DETR can be divided into five parts: Unimodal Encoders, Distill Align, Convolutional Fuser, Loop Decoder, and Prediction Heads. 

The input video and text are firstly fed into pre-trained feature extractors to extract video and text features $V\in\mathbb{R}^{b \times n \times d_v}, T\in\mathbb{R}^{b \times n \times d_t}$, where $b$ is the batch size, $d_v$ and $d_t$ are the dimensions of the extracted video clip and text token features respectively. Then the video and text features are fed into two unimodal encoders $\text{UE}_{v}(\cdot), \text{UE}_{t}(\cdot)$ to be mapped into latent spaces $V'\in\mathbb{R}^{b \times t \times d}, T'\in\mathbb{R}^{b \times n \times d}$, where $d$ is the hidden dimension of the model. We use the two unimodal encoders and two momentum unimodal encoders $\text{UE}_{vm}(\cdot), \text{UE}_{tm}(\cdot)$ to obtain global features. The two global features and two momentum global features $G_v, G_{mv}, G_t, G_{mt}\in\mathbb{R}^{b \times d}$ obtained by unimodal encoders are fed into a Distill Align method to ensure that the features are mapped to the same space. Then the mapped features $V', T'$ are sent to the Convolutional Fuser to obtain multimodal features $R\in\mathbb{R}^{b \times t \times d}$. Then the multimodal features $R$ are sent into the Loop Decoder together with a zero matrix $O\in\mathbb{R}^{b \times q \times d}$ to obtain the decoded features $R'\in\mathbb{R}^{b \times q \times d}$, where $q$ is a hyperparameter representing the number of reference point. Finally, the decoded features are sent to the prediction heads to obtain predicted moments $\{ c_{\textnormal{m}}, w_{\textnormal{m}} \}_{\textnormal{m}}^m$ and saliency scores $s\in\mathbb{R}^t$.

\subsection{Unimodal Encoders and Distill Align}
\label{subsec:unienc}

We use a two-layer multilayer perceptron~\cite{rumelhart1986learning} as a unimodal encoder $\text{UE}(\cdot)$ to map the extracted features $X\in\mathbb{R}^{b \times x \times d_x}$ to the latent space $X'\in\mathbb{R}^{b \times x \times d}$, 
% \begin{align}
%     X'_{\rb, \rx} &= \text{UE}( X_{\rb, \rx} ),  
% \end{align}
where $X\in \{ V, T \}$ and $x\in \{ t, n \}$. 
We use the average of the features of the sample in the latent space $X'$ in the clip or token dimension to obtain the global features $G\in\mathbb{R}^{b \times d}$ of each sample. 

% To keep the consistency of the extracted global features, 
We use two learnable unimodal encoders $\text{UE}(\cdot)\in\{ \text{UE}_{v}(\cdot), \text{UE}_{t}(\cdot) \}$ and two momentum unimodal encoders $\text{UE}_{m}(\cdot)\in\{ \text{UE}_{mv}(\cdot), \text{UE}_{mt}(\cdot) \}$ to map the extracted features $V\in\mathbb{R}^{b \times x \times d_v}, T\in\mathbb{R}^{b \times x \times d_t}$ into a latent space, and obtaine four global features $G_v, G_{mv}, G_t, G_{mt}$ respectively. Momentum unimodal encoders are updated from the corresponding unimodal encoders: 
\begin{align}
    \text{UE}_{m\theta}^0 &= \text{UE}_{\theta}^0, \\
    \text{UE}_{m\theta}^i &= m \text{UE}_{m\theta}^{i-1} + (1-m)\text{UE}_{\theta}^i,&\text{when } i > 0, 
\end{align}
where $m \in [0, 1)$ is a momentum coefficient, $\text{X}_{\theta}$ means all parameters in the method $\text{X}$. Obtained the mapped features $V', V_{m}', T', T_{m}'\in\mathbb{R}^{b \times x \times d}$, we calculate the global features for each sample $G_v, G_{mv}, G_t, G_{mt}\in\mathbb{R}^{b \times d}$. Following MoCo~\cite{he2020momentum}, we push the momentum global features into momentum global features queues $Q_v, Q_t\in\mathbb{R}^{l \times d}$, where $l$ represents the queue length. 

\begin{figure}[!h]
\centering
\includegraphics[width=0.6\textwidth]{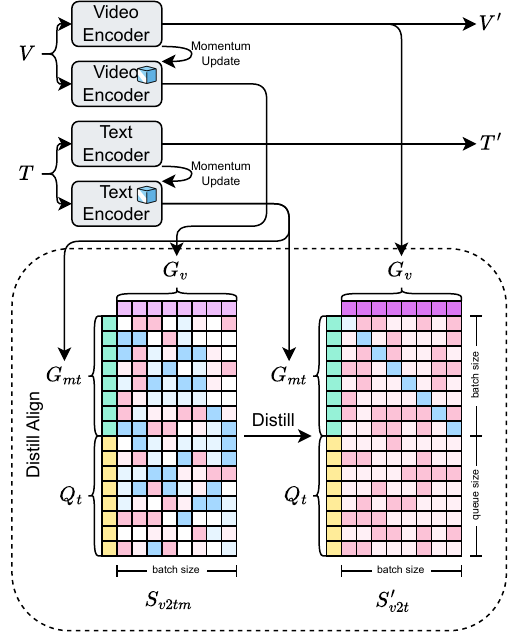}
\caption{\textbf{The structure of the Distill Align. }}
\label{fig:queue_align}
\end{figure}

Figure \ref{fig:queue_align} shows the structure of the Distill Align. At this point, we have video global features $G_{v}\in\mathbb{R}^{b \times d}$, video and text momentum global features $G_{mv}, G_{mt}\in\mathbb{R}^{b \times d}$, and text momentum global features queues $Q_{t}\in\mathbb{R}^{ l \times d}$. By calculating the cosine similarity $\text{S}(\cdot, \cdot)$ we get the video to text similarity matrix $S_{v2t}, S_{v2tm}\in\mathbb{R}^{b \times l}$ among these global features: 
\begin{align}
    S_{v2t} &= \text{S}(G_{v}, Q_{t}), \\
    S_{v2tm} &= \text{S}(G_{mv}, Q_{t}).
\end{align}
To mitigate the impact of overlapping semantic information, following ALBEF~\cite{moon2023correlation} we distill the similarity matrix $S_{v2tm}\in\mathbb{R}^{b \times l}$ into an identity matrix $I$ as the goal $S'_{v2t}\in\mathbb{R}^{b \times l}$ of multimodal alignment: 
\begin{align}
    S'_{v2t} &= \alpha S_{v2tm} + (1-\alpha) I, 
\end{align}
where $\alpha\in [0, 1)$ is the distillation coefficient. Finally, the video-to-text alignment loss $\mathcal{L}_{v2t}$ is
\begin{align}
    \mathcal{L}_{v2t} &= \text{CE}(S_{v2t}, S'_{v2t}), 
\end{align}
where $\text{CE}(\cdot, \cdot)$ is cross entropy loss. Following SimSiam~\cite{chen2021exploring}, the video-to-text similar loss is
\begin{align}
    \mathcal{L}_{v2tsim} &= - \text{mean}(\text{s}(G_{v}, \text{Linear}(Q_{t}))), 
\end{align}
where $\text{mean}(\cdot)$ is the average of the matrix, $\text{Linear}(\cdot)$ is a linear layer. 

Similarly, we calculate the text-to-video alignment loss $\mathcal{L}_{t2v}$ and the text-to-video similar loss $\mathcal{L}_{t2vsim}$. The final multimodal alignment loss is
\begin{align}
    \mathcal{L}_{align} = \lambda_{align} (\mathcal{L}_{v2t} + \mathcal{L}_{t2v}) / 2 + \lambda_{sim} (\mathcal{L}_{v2tsim} + \mathcal{L}_{t2vsim}) / 2.
\end{align}

\subsection{Convolutional Fuser}

% Current models only focus on removing features that are not related to text in video features, while ignoring the temporal structure of the video itself. They extract video features with Transformer Encoder~\cite{vaswani2017attention}, whose global attention calculation~\cite{bahdanau2014neural} may dilute the weight of local details, causing the model to ignore the extraction of local features of the video. In image recognition tasks where local features are equally important, the local receptive field of the convolutional layer~\cite{krizhevsky2012imagenet, zeiler2014visualizing} allows the network to capture local shapes and patterns, thereby improving the performance of the model. Inspired by this, we introduced 1D convolutional layer in Multimodal Fusion and Extraction Method. 

\begin{figure}[!h]
\centering
\includegraphics[width=0.6\textwidth]{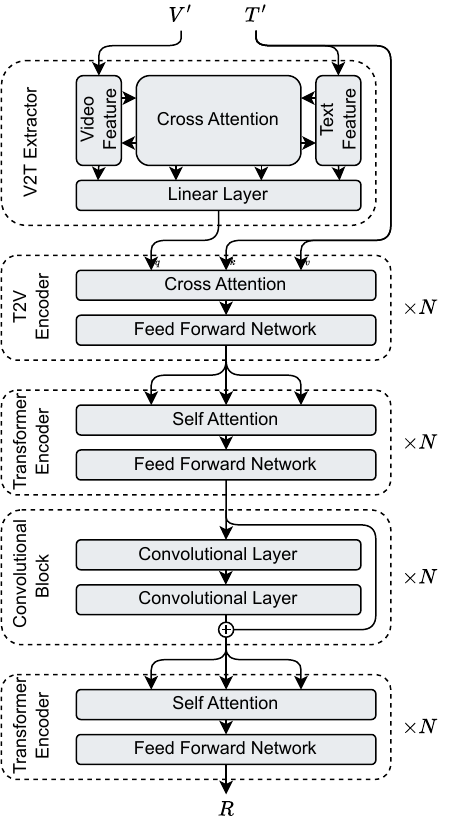}
\caption{\textbf{The structure of the Convolutional Fuser. }}
\label{fig:1d_conv_method}
\end{figure}

Figure \ref{fig:1d_conv_method} shows the structure of the Convolutional Fuser. 
First, we feed the video features $V'\in\mathbb{R}^{b \times t \times d}$ and text features $T'\in\mathbb{R}^{b \times n \times d}$ mapped to the latent space into a V2T Extractor~\cite{sun2024tr}, a T2V Encoder~\cite{moon2023query} and a Transformer Encoder~\cite{vaswani2017attention} to obtain text-irrelevant video features $V''\in\mathbb{R}^{b \times t \times d}$. 
Subsequently, the text-irrelevant video features $V''\in\mathbb{R}^{b \times t \times d}$ are fed into convolutional blocks. After passing through this residual network, we obtain local enhanced multimodal fusion features $V'''\in\mathbb{R}^{b \times t \times d}$. 
Finally, the locally enhanced multimodal fusion features $V'''\in\mathbb{R}^{b \times t \times d}$ are fed into another Transformer Encoder~\cite{vaswani2017attention} to obtain multimodal features $R\in\mathbb{R}^{b \times t \times d}$. 

\subsubsection{V2T Extractor}

First, following TR-DETR~\cite{sun2024tr}, we use the video features $V'\in\mathbb{R}^{b \times t \times d}$ and text features $T'\in\mathbb{R}^{b \times n \times d}$ which have been mapped to the latent space to obtain the correlation matrix $A\in\mathbb{R}^{b \times t \times n}$ between video features and text features: 
\begin{align}
    A_1 &= \text{Linear}(V'), \\
    A_2 &= \text{Linear}(T'), \\
    A_3 &= \text{Linear}(V') T'^T, \\
    A &= A_1 + A_2^T + A_3, 
\end{align}
when matrices are added, the shorter dimension expands itself to the same length as the other matrices. Then we perform softmax on the correlation matrix $A$ in the text and video dimensions to obtain two other correlation matrices $A_r, A_c\in\mathbb{R}^{b \times t \times n}$. Then we get text-irrelevant video features $V'_v\in\mathbb{R}^{b \times t \times d}$: 
\begin{align}
    T_{v} &= A_r T', \\
    V_{t} &= A_t A_c^T V', \\
    V_{cat} &= \left[ V' || T_{v} || V' \circ T_{v} || V' \circ V_{t} \right], \\
    V'_{cat} &= \text{Linear}(V_{cat}), \\
    B &= \text{Softmax}(\text{Linear}(T')), \\
    T_p &= T'^T B, \\
    V''_{cat} &= \left[ V'_{cat} || T_p \right], \\
    V'_v &= \text{Linear}(V''_{cat}), 
\end{align}
where $[ \cdot || \cdot ]$ represents concatenation, $\circ$ represents the Hadamard product. 

\subsubsection{T2V Encoder}

Then, following QD-DETR~\cite{moon2023query}, we use text-irrelevant video features $V'_v$ to obtain text-guided text-irrelevant video features $V''_v\in\mathbb{R}^{b \times t \times d}$:
\begin{align}
    Q_v &= \text{Linear}(V'_v), \\
    K_t &= \text{Linear}(T'), \\
    V_t &= \text{Linear}(T'), \\
    \text{Attention}(Q_v, K_t, V_t) &= \text{Softmax}\left( \dfrac{Q_v K_t^T}{d} \right) V_t, \\
    V''_v &= \text{FFN}(\text{Attention}(Q_v, K_t, V_t)). 
\end{align}

\subsubsection{Transformer Encoder 1}

The text-guided text-irrelevant video features $V''_v\in\mathbb{R}^{b \times t \times d}$ are then fed into a Transformer Encoder to obtain the text-irrelevant video features $V''\in\mathbb{R}^{b \times t \times d}$. The Transformer Encoder here is no different from that in other papers~\cite{vaswani2017attention}. 

\subsubsection{Convolutional Blocks}

Subsequently, the text-irrelevant video features $V''\in\mathbb{R}^{b \times t \times d}$ are fed into convolutional blocks. Following previous work in image recognition, we use residual blocks $\text{RB}(\cdot)$ similar to ResNet~\cite{he2016deep}: 
\begin{align}
    X_{i} &= \text{RB}_{i}(X_{i-1}) = \sigma(X_{i-1} + \text{BN}(\text{Conv}(\sigma(\text{BN}(\text{Conv}(X_{i-1})))))), \label{eq:residual_block}
    % X_{i} &= \text{RB}_{i}(X_{i-1}) = \sigma(X_{i-1} + \mathcal{F}(X_{i-1})), \label{eq:residual_block} \\
    % \mathcal{F}(X_{i-1}) &= \text{BN}(\text{Conv}(\sigma(\text{BN}(\text{Conv}(X_{i-1}))))), 
\end{align}
where $\text{Conv}(\cdot)$ represents an Convolutional Layer, $\text{BN}(\cdot)$ represents the batch normalization, and $\sigma(\cdot)$ represents an activation function. We stack $N$ residual blocks to extract local information in video features: 
\begin{align}
    X_0 &= V'', \\
    X_i &= \text{RB}_i(X_{i-1}),& \text{when } i > 0, \tag{\ref{eq:residual_block}}\\
    V''' &= X_{N}.
\end{align}
After passing through this residual network, we obtain local enhanced multimodal fusion features $V'''\in\mathbb{R}^{b \times t \times d}$. 

\subsubsection{Transformer Encoder 2}

The local enhanced multimodal fusion features $V'''\in\mathbb{R}^{b \times t \times d}$ are then fed into another Transformer Encoder to obtain multimodal features $R\in\mathbb{R}^{b \times t \times d}$. The Transformer Encoder here is also no different from that in other papers~\cite{vaswani2017attention}. It should be noted that the two Transformer Encoders here do not share parameters.

\subsection{Loop Decoder}

Research shows that when the Transformer Decoder’s $query$ carries target information, the decoder can better decode the input information~\cite{liu2022dab}, and the output of the Transformer Decoder carries exactly the target information. Inspired by this, we feed the Transformer Decoder's output back into itself as the $query$ to enhance its ability to decode multimodal fusion information. 

\begin{figure}[!h]
\centering
\includegraphics[width=0.6\textwidth]{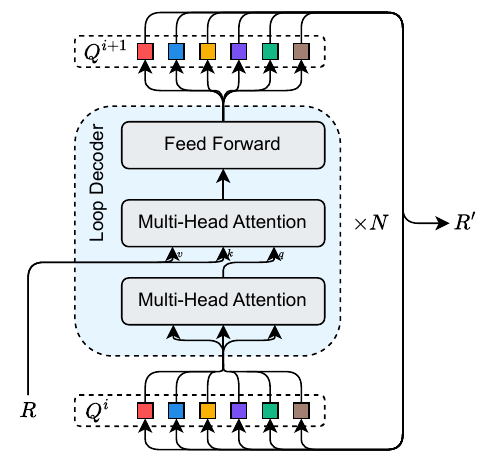}
\caption{\textbf{The structure of the Loop Decoder. }}
\label{fig:loop_decoder}
\end{figure}

Figure \ref{fig:queue_align} shows the structure of the Loop Decoder. We feed a zero matrix $O\in\mathbb{R}^{b \times q \times d}$ and multimodal features $R\in\mathbb{R}^{b \times t \times d}$ into the Transformer Decoder $\text{TD}(\cdot, \cdot)$~\cite{vaswani2017attention}. Decoded features $R'\in\mathbb{R}^{b \times q \times d}$ are obtained after $N$ loops of Transformer Decoder: 
\begin{align}
    Q^0 &= O, \\
    Q^i &= \text{TD}(Q^{i-1}, R),&\text{when } i > 0, \\
    R' &= Q^{N},
\end{align}
where $q$ is a hyperparameter representing the number of reference points. Compared with a bigger decoder, Loop Decoder does not introduce new parameters and does not cause overfitting.

\subsection{Prediction Heads}

% We use prediction heads similar to TR-DETR~\cite{sun2024tr}. 
% The structure of Prediction Heads is introduced in Section \ref{sec:prediction_heads}. 

We use prediction heads similar to QD-DETR~\cite{moon2023query} and TR-DETR~\cite{sun2024tr}. 

\subsubsection{Video Moment Retrieval Prediction Head}

We feed the decoded features $R'\in\mathbb{R}^{b \times q \times d}$ into a multi-layer perceptron to obtain $q$ predicted moments $\{ c_{\textnormal{m}}, w_{\textnormal{m}} \}_{\textnormal{m}}^q$:
\begin{align}
    \{s, e\} &= \text{MLP}(R'),
\end{align}
where $s$ and $e$ represents the start and end of one predicted moments. At the same time, the decoded features $R$ and multimodal features $R'\in\mathbb{R}^{b \times t \times d}$ are used to obtain the confidence of each predicted moments $p\in\mathbb{R}^q$:
\begin{align}
    p &= \text{MLP}(R') + \text{Sigmoid}^{-1}(\text{MLP}(R')),
\end{align}
where $\text{Sigmoid}^{-1}(\cdot)$ represents reverse sigmoid. In this way, we get all moments $\{ c_{\textnormal{m}}, w_{\textnormal{m}} \}_{\textnormal{m}}^m$ related to the text query in the video. 

\subsubsection{Highlight Detection Prediction Head}

After getting all the moments, we take out the decoded features corresponding to all the clips in the moments $R''\in\mathbb{R}^{b \times t' \times d}$, where $t'$ represents the amount of total clips in the Video Moment Retrieval prediction, feed them into a Gated Recurrent Unit~\cite{chung2014empirical}, and use the hidden state as the new global features of the video $G'_v\in\mathbb{R}^{b \times d}$: 
\begin{align}
    o =& \text{GRU}(R''), \\
    G'_v =& \text{GRU}_{\theta},
\end{align}
where $o$ represents the output of the Gated Recurrent Unit which we do not care about, $\text{GRU}_{\theta}$ represents the Gated Recurrent Unit's hidden state. Then, we calculate the similarity between the new global features $G'_v$ and the features $V'$ of each clip:
\begin{align}
    S &= G'_v V'^T
\end{align}
and finally get saliency scores $s\in\mathbb{R}^t$: 
\begin{align}
    M &= \text{Linear}(R' \circ S + R'), \\
    s &= \text{sum}(M) / d,
\end{align}
where $\circ$ represents the Hadamard product, $\text{sum}(\cdot)$ means to sum matrix elements over columns, $d$ is the hidden dimension of the model. 

\subsection{Objective Losses}

% We use objective losse similar to QD-DETR~\cite{moon2023query}. 
The objective loss function of LD-DETR $\mathcal{L}_{total}$ is
\begin{align}
    \mathcal{L}_{total} &= \mathcal{L}_{mr} + \mathcal{L}_{hd} + \mathcal{L}_{align},
\end{align}
where 
\begin{align}
    % \mathcal{L}_{mr} &= \mathcal{L}_{mom} + \lambda_{CE} \text{CE}(\hat{y}, y), \\
    % \mathcal{L}_{mom} &= \lambda_{L1} ||m-\hat{m}|| + \lambda_{gIoU} \mathcal{L}_{gIoU}(m, \hat{m}), \\
    % \mathcal{L}_{hd} &= \lambda_{marg} \mathcal{L}_{marg} + \lambda_{cont} \text{RAC}(X_r^{\text{pos}}, X_r^{\text{neg})}) \\
    % \mathcal{L}_{marg} &= \text{max}(0, \Delta + S(x^{\text{low}}) - S(x^{\text{high}})), 
    \mathcal{L}_{mr} &= \lambda_{L1} ||m-\hat{m}|| + \lambda_{gIoU} \mathcal{L}_{gIoU}(m, \hat{m}) + \lambda_{CE} \text{CE}(\hat{y}, y), \\
    \mathcal{L}_{hd} &= \lambda_{marg} \text{max}(0, \Delta + S(x^{\text{low}}) - S(x^{\text{high}})) + \lambda_{cont} \text{RAC}(X_r^{\text{pos}}, X_r^{\text{neg})}), 
    % \mathcal{L}_{cont} &= -\sum_{r=1}^{R}\log \frac{\sum_{x\in X_r^{\text{pos}}}\exp(S(x)/\tau}{\sum_{x\in (X_r^{\text{pos}} \cup X_r^{\text{neg})}}\exp(S(x)/\tau},
\end{align}
where $y$ and $\hat{y}$ are the ground-truth of either to foreground or background and its correspond prediction, $m$ and $\hat{m}$ are ground-truth moment and its correspond prediction, IoU loss $\mathcal{L}_{gIoU}(\cdot, \cdot)$ is from a previous work~\cite{rezatofighi2019generalized}, $\Delta$ is the margin, $S(\cdot)$ is the saliency score estimator, $x^{\text{high}}$ and $x^{\text{low}}$ are video tokens from two pairs of high and low-rank clips respectively, $\text{CE}(\cdot, \cdot)$ is cross entropy loss, $\text{RAC}(\cdot, \cdot)$ is the rank-aware contrastive loss~\cite{hoffmann2022ranking}, $X_r^{\text{pos}}$ and $X_r^{\text{neg}}$ are the set of samples with a higher and lower rank than the iteration index respectively. 

\clearpage

\section{Experiments}

\subsection{Datasets}

% \subsubsection{Datasets}

We evaluated our model on four prevalent Video Moment Retrieval and Highlight Detection public benchmarks: QVHighlights~\cite{lei2021detecting}, Charade-STA~\cite{gao2017tall}, TACoS~\cite{regneri2013grounding}, and TVSum~\cite{song2015tvsum}. 

Since the dataset limits the number of submissions, in Comparison with Models, we conducted multiple experiments and gave the result of the experiment that was most likely to get the best result. 

Because QVHighlight~\cite{lei2021detecting} is the only existing data set that supports both Video Moment Retrieval and Highlight Detection. In Ablation Studies, we conducted all experiments on the QVHighlight dataset \textit{val} split. We conducted each experiment five times, using 1, 23, 456, 7,890 and 12,345 as random seeds respectively, and the average and variance of all experimental results are given.

\subsection{Metrics}

We adopted the same evaluation metrics with previous works. To be specific, we computed Recall@1 with IoU threshold $\theta_{IoU} = 0.5 \text{ and } 0.7$, mean average precision (mAP) with $\theta_{IoU} = 0.5 \text{ and } 0.7$, and mAP with a series of thresholds $[0.5:0.05:0.95]$ for Video Moment Retrieval on QVHighlights. mAP and HIT@1 where positive samples were defined as with the saliency score of $v$ were adopted for Highlight Detection. On Charades-STA and TACoS datasets, we utilized Recall@1 with $\theta_{IoU} = \{0.3, 0.5, 0.7\}$ and mIoU to measure the Video Moment Retrieval performance. For TVSum, mAP and Top-5 mAP are adopted, respectively.

\subsection{Experimental Settings}

In all experiments, we adopted CLIP~\cite{radford2021learning} and Slowfast~\cite{feichtenhofer2019slowfast} as extractors to extract video features, and adopted CLIP~\cite{radford2021learning} to extract text features. In some experiments, adopted PANN~\cite{kong2020panns} to extract audio features. 

By default, our model trained 250 epochs with AdamW optimizer~\cite{loshchilov2017decoupled} using a learning rate of 1e-4, a batch size of 32, a hidden dimension of 256, a queue length of 65,536, a momentum coefficient of 0.995, a distillation coefficient of 0.4, a align coefficient $\lambda_{align}$ of 0.6, a similar coefficient $\lambda_{sim}$ of 0.6, 5 layers of Convolutional Blocks, 10 number of reference points, and 3 loops of the Loop Decoder. 
In the TACoS dataset, we used a distillation coefficient of 0.7. 
In the Charade-STA dataset, we trained 100 epochs using a distillation coefficient of 0.3, a similar coefficient $\lambda_{sim}$ of 0.4, and 4 layers of Convolutional Blocks. 
In the TVSum dataset, 
% we tried every method and changed every hyperparameter to achieve a better result. Because of this dataset's characteristics, the hyperparameters in each experiment are different. So, 
we did not record them. 

\subsection{Comparison with Other Models}

\begin{figure*}[!h]
\centering
\includegraphics[width=1\textwidth]{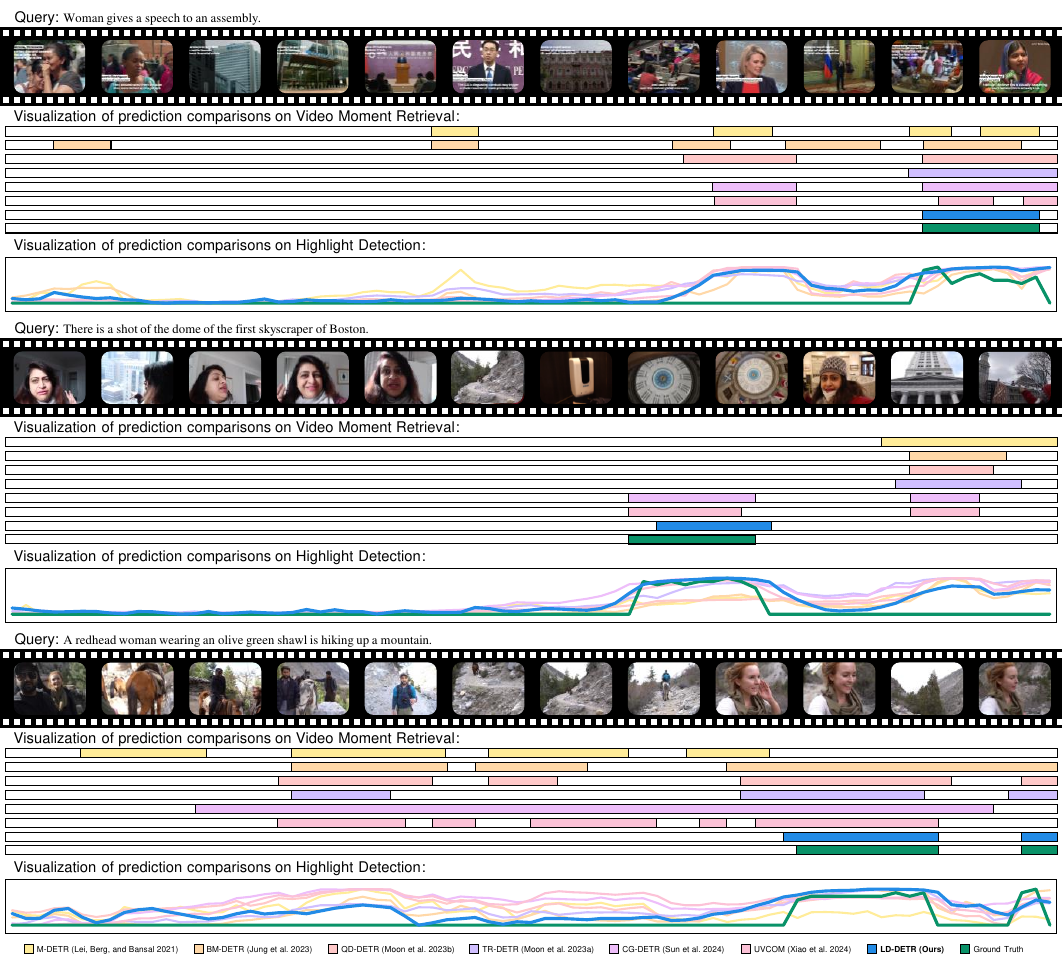}
\caption{\textbf{Visualization comparison on Video Moment Retrieval and Highlight Detection. }}
\label{fig:visualization}
\end{figure*}

Figure \ref{fig:visualization} visualizes comparison on LD-DETR and other models. 
Table \ref{tab:qvhighlights_experiment} reports the LD-DETR's performance on joint Video Moment Retrieval and Highlight Detection tasks on QVHighlight dataset. All of the models given in the table don't use pre-training. The models in the table are divided into three categories according to the extractors used. Our model outperforms all existing models, even those using more extracted features. Table \ref{tab:charades_tacos_experiment} reports the LD-DETR's performance on Video Moment Retrieval on TACoS dataset and Charades-STA dataset. 
Benefiting from our proposed method, the LD-DETR model outperforms the State-Of-The-Art models on QVHighlight, Charades-STA and TACoS datasets. 

\begingroup
\fontsize{9}{11}\selectfont
\begin{table*}[!h]
    \centering
    \scalebox{0.77}{
    \setlength{\tabcolsep}{4mm}
        \begin{tabular}{lccccccc}
        \toprule[0.15em]
        \multirow{3}{*}{\textbf{Model}} & \multicolumn{5}{c}{\textbf{Video Moment Retrieval}} & \multicolumn{2}{c}{\textbf{Highlight Detection}} \\
        \cmidrule(l){2-6}  \cmidrule(l){7-8}
        & \multicolumn{2}{c}{R1} & \multicolumn{3}{c}{mAP} & \multicolumn{2}{c}{$>=$Very Good} \\
        \cmidrule(l){2-3} \cmidrule(l){4-6}  \cmidrule(l){7-8}
        & @0.5 & @0.7 & @0.5 & @0.75 & Avg. & mAP & HIT@1 \\

        \midrule[0.15em] 
        \multicolumn{8}{l}{\textbf{Slowfast} $+$ \textbf{CLIP} \textsubscript{(5.5 GB)} } \\
        \midrule
        BeautyThumb~\cite{song2016click} & - & - & - & - & - & 14.36 & 22.88 \\
        DVSE~\cite{liu2015multi} & - & - & - & - & - & 18.75 & 21.79 \\
        MCN~\cite{anne2017localizing} & 11.41 & 2.72 & 24.94 & 8.22 & 10.67 & - & - \\
        CAL~\cite{escorcia2019finding} & 25.49 & 11.54 & 23.40 & 7.65 & 9.89 & - & - \\
        XML~\cite{lei2020tvr} & 41.83 & 30.35 & 44.63 & 31.73 & 32.14 & 34.49 & 55.25 \\
        XML\textsuperscript{+}~\cite{lei2021detecting} & 46.69 & 33.46 & 47.89 & 34.67 & 34.90 & 35.38 & 55.06 \\
        Moment-DETR~\cite{lei2021detecting} & 52.89 & 33.02 & 54.82 & 29.40 & 30.73 & 35.69 & 55.60 \\
        Localizer~\cite{yu2024self} & 54.50 & 36.50 & - & - & 32.30 & - & - \\
        UniVTG~\cite{lin2023univtg} & 58.86 & 40.86 & 57.60 & 35.59 & 35.47 & 38.20 & 60.96 \\
        MomentDiff~\cite{li2024momentdiff} & 57.42 & 39.66 & 54.02 & 35.73 & 35.95 & - & - \\
        VMRNet~\cite{JSYJ20240520004} & 59.94 & 42.84 & 55.56 & 37.75 & 36.87 & - & - \\
        LLaViLo~\cite{ma2023llavilo} & 59.23 & 41.42 & 9.72 & - & 36.94 & - & - \\
        MH-DETR~\cite{xu2024mh} & 60.05 & 42.48 & 60.75 & 38.13 & 38.38 & 38.22 & 60.51 \\
        QD-DETR~\cite{moon2023query} & 62.40 & 44.98 & 62.52 & 39.88 & 39.86 & 38.94 & 62.40 \\
        CDIM~\cite{1024497142.nh} & 60.51 & 45.53 & 61.36 & 41.05 & 39.94 & 37.69 & 60.05 \\
        BM-DETR~\cite{jung2023overcoming} & 60.12 & 43.05 & 63.08 & 40.18 & 40.08 & - & - \\
        MESM~\cite{liu2024towards} & 62.78 & 45.20 & 62.64 & 41.45 & 40.68 & - & - \\
        \eval{EaTR~\cite{jang2023knowing}} & \eval{61.36} & \eval{45.79} & \eval{61.86} & \eval{41.91} & \eval{41.74} & \eval{37.15} & \eval{58.65} \\
        LMR~\cite{liu2024context} & 64.40 & 47.21 & 64.65 & 43.16 & 42.56 & - & - \\
        TR-DETR~\cite{sun2024tr} & 64.66 & 48.96 & 63.98 & 43.73 & 42.62 & 39.91 & 63.42 \\
        CG-DETR~\cite{moon2023correlation} & 65.43 & 48.38 & 64.51 & 42.77 & 42.86 & \second{40.33} & \best{66.21} \\
        \eval{CDNet~\cite{ma2024disentangle}} & \eval{67.74} & \eval{49.55} & \eval{63.82} & \eval{42.30} & \eval{42.76} & \eval{39.84} & \eval{66.52} \\
        UVCOM~\cite{xiao2024bridging} & 63.55 & 47.47 & 63.37 & 42.67 & 43.18 & 39.74 & 64.20 \\
        SFABD~\cite{huang2024semantic} & - & - & 62.38 & 44.39 & 43.79 & - & - \\
        LLMEPET~\cite{jiang2024prior} & \second{66.73} & 49.94 & \second{65.76} & 43.91 & 44.05 & \second{40.33} & \second{65.69} \\
        UniVTG\textsuperscript{+}~\cite{chen2024video} & 66.65 & \best{52.19} & 64.37 & \second{46.68} & 45.18 & 40.18 & 64.77 \\
        BAM-DETR~\cite{lee2023bam} & 62.71 & 48.64 & 64.57 & 46.33 & \second{45.36} & - & - \\
        \eval{TaskWeave~\cite{yang2024task}} & \eval{64.26} & \eval{50.06} & \eval{65.39} & \eval{46.47} & \eval{45.38} & \eval{39.28} & \eval{63.68} \\
        
        % \rowcolor{my_blue}
        \textbf{LD-DETR (Ours)} & \best{66.80} & \second{51.04} & \best{67.61} & \best{46.99} & \best{46.41} & \best{40.51} & 65.11 \\
        % MCN &  &  &  &  &  &  &  \\

        \midrule
        \multicolumn{8}{l}{\textbf{Slowfast} $+$ \textbf{CLIP} $+$ \textbf{PANN} \textsubscript{(11.7 GB)} } \\
        \midrule
        UMT~\cite{liu2022umt} & 56.23 & 41.18 & 53.38 & 37.01 & 36.12 & 38.18 & 59.99 \\
        VCSJT~\cite{zhou2024subtask} & 59.14 & 42.02 & 55.76 & 37.79 & 36.37 & 38.59 & 62.45 \\
        MIM~\cite{li2023mim} & 59.99 & 41.50 & 55.85 & 36.84 & 36.45 & 38.96 & 62.39 \\
        LSJT~\cite{wang2024modality} & 60.51 & 41.50 & 56.33 & 36.70 & 36.66 & 39.13 & 61.22 \\
        MomentDiff~\cite{li2024momentdiff} & 58.21 & 41.48 & 54.57 & 37.21 & 36.84 & - & - \\
        QD-DETR~\cite{moon2023query} & 63.06 & 45.10 & 63.04 & 40.10 & 40.19 & 39.04 & 62.87 \\
        UVCOM~\cite{xiao2024bridging} & \second{63.81} & \second{48.70} & \second{64.47} & \second{44.01} & \second{43.27} & \second{39.79} & \second{64.79} \\
        % BAM-DETR~\cite{lee2023bam} & 64.07 & 48.12 & 65.61 & 47.51 & 46.91 & - & - \\
        
        % \rowcolor{my_blue}
        \textbf{LD-DETR (Ours)} & \best{65.76} & \best{50.71} & \best{66.06} & \best{46.62} & \best{45.85} & \best{41.00} & \best{67.06} \\

        \midrule
        \multicolumn{8}{l}{\textbf{CLIP\textsuperscript{+}} \textsubscript{(233.7 GB)} } \\
        \midrule
        R\textsuperscript{2}-Tuning~\cite{liu2024r} & 68.03 & 49.35 & 69.04 & 47.56 & 46.17 & 40.75 & 64.20 \\
        
        \bottomrule[0.15em]
    \end{tabular}{}
    }
    \caption{\textbf{Jointly Video Moment Retrieval and Highlight Detection results on QVHighlights \textit{test} split. }The table categorizes the models by the used extractors, and indicates the extractors and the extracted features sizes. The best result in each category of features on \textit{test} split in each column is highlighted in \best{bold}, and the second best result is highlighted in \second{underline}. The models shown in \eval{italic} only report their results on the \textit{val} split in the paper. 
    }
    \label{tab:qvhighlights_experiment}
\end{table*}
\endgroup

\begingroup
\fontsize{9}{11}\selectfont
\begin{table*}[!h]
    \centering
    \scalebox{0.71}{
    \setlength{\tabcolsep}{2.3mm}
        \begin{tabular}{lcccccccc}
        \toprule[0.15em]
        \multirow{2}{*}{\textbf{Model}} & \multicolumn{4}{c}{\textbf{TACoS}} & \multicolumn{4}{c}{\textbf{Charades-STA}} \\
        \cmidrule(l){2-5} \cmidrule(l){6-9} & R1@0.3 & R1@0.5 & R1@0.7 & mIoU & R1@0.3 & R1@0.5 & R1@0.7 & mIoU \\

        \midrule[0.15em] 

        CTRL~\cite{gao2017tall} & 18.32 & 13.30 & - & - & - & 23.63 & 8.89 & - \\
        ABLR~\cite{yuan2019find} & 19.50 & 9.40 & - & 13.40 & - & - & - & - \\
        SM-RL~\cite{wang2019language} & - &  20.25 & 15.95 & - & - & 24.36 & 11.17 & - \\
        TGN~\cite{chen2018temporally} & 21.77 & 18.90 & - & - & - & - & - & - \\
        SAP~\cite{chen2019semantic} & - & 18.24 & - & - & - & 27.42 & 13.36 & - \\
        MIM$\dagger$~\cite{li2023mim} & - & - & - & - & - & 43.92 & 25.89 & - \\
        MAN~\cite{zhang2019man} & - & - & - & - & - & 46.53 & 22.72 & - \\
        FMAN~\cite{BJHK20231124001} & - & - & - & - & - & 51.40 & 25.05 & 38.23 \\
        LSJT$\dagger$~\cite{wang2024modality} & - & - & - & - & - & 44.62 & 25.13 & - \\
        DRN~\cite{zeng2020dense} & - & 23.17 & - & - & - & 45.40 & 26.40 & - \\
        UMT$\dagger$~\cite{liu2022umt} & - & - & - & - & - & 48.31 & 29.25 & - \\
        VCSJT$\dagger$~\cite{zhou2024subtask} & - & - & - & - & - & 51.21 & 30.22 & - \\
        SFABD~\cite{huang2024semantic} & - & - & - & - & - & - & 30.51 & - \\
        BPNet~\cite{xiao2021boundary} & 25.96 & 20.96 & 14.08 & 19.53 & 65.48 & 50.75 & 31.64 & 46.34 \\
        Moment-DETR~\cite{lei2021detecting} & - & - & - & - & - & 53.63 & 31.37 & - \\
        SCDM~\cite{yuan2019semantic} & 26.11 & 21.17 & - & - & - & 54.44 & 31.37 & - \\
        SCDN~\cite{PZKX202208003} & 27.64 & 23.27 & - & - & - & 54.92 & 34.26 & - \\
        DCL~\cite{nan2021interventional} & 38.84 & 29.07 & 19.05 & 28.26 & 67.63 & 50.24 & 32.88 & 48.02 \\
        HUAL~\cite{ji2023binary} & - & - & - & - & 70.40 & 52.69 & 28.90 & 48.11 \\
        VSLNet~\cite{zhang2020span} & 29.61 & 24.27 & 20.03 & 24.11 & 70.46 & 54.19 & 35.22 & 50.02 \\
        2D-TAN~\cite{zhang2020learning} & 37.29 & 25.32 & - & - & - & 39.70 & 23.31 & - \\
        MMN~\cite{wang2022negative} & 39.24 & 26.17 & - & - & - & 47.31 & 27.28 & - \\
        CrossGraphAlign~\cite{PZKX202006007} & 39.80 & 26.40 & - & - & - & - & - & - \\
        CBLN~\cite{liu2021context} & 38.98 & 27.65 & - & - & - & 61.13 & 38.22 & - \\
        CPNet~\cite{li2021proposal} & 42.61 & 28.29 & - & 28.69 & - & 60.27 & 38.74 & - \\
        FVMR~\cite{gao2021fast} & 41.48 & 29.12 & - & - & - & 59.46 &  35.48 & - \\
        SimVTP~\cite{ma2022simvtp} & 43.10 & 30.30 & - & - & - & 44.70 & 26.30 & - \\
        RaNet~\cite{gao2021relation} & 43.34 & 33.54 & - & - & - & 60.40 & 39.65 & - \\
        MomentDiff~\cite{li2024momentdiff} & 44.78 & 33.68 & - & - & - & 55.57 & 32.42 & - \\
        LLaViLo~\cite{ma2023llavilo} & - & - & - & - & - & 55.72 & 33.43 & - \\
        TaskWeave~\cite{yang2024task} & - & - & - & - & - & 56.51 & 33.66 & - \\
        QD-DETR$\dagger$~\cite{moon2023query} & - & - & - & - & - & 55.51 & 34.17 & - \\
        LMR~\cite{liu2024context} & - & - & - & - & - & 55.91 & 35.19 & - \\
        VLG-Net~\cite{soldan2021vlg} & 45.46 & 34.19 & - & - & - & - & - & - \\
        GVL~\cite{wang2023learning} & 45.92 & 34.57 & - & 32.48 & - & - & - & - \\
        TR-DETR~\cite{sun2024tr} & - & - & - & - & - & 57.61 & 33.52 & - \\
        UniVTG~\cite{lin2023univtg} & 51.44 & 34.97 & 17.35 & 33.60 & 70.81 & 58.01 & 35.65 & 50.10 \\
        UniVTG\textsuperscript{+}~\cite{chen2024video} & - & - & - & - & 68.06 & 57.18 & 36.05 & - \\
        BM-DETR~\cite{jung2023overcoming} & 50.31 & 35.42 & - & - & - & 59.48 & 38.33 & - \\
        SFEN~\cite{DZYX202212037} & 47.30 & 36.10 & - & - & - & - & - & - \\
        % D-TSG~\cite{liu2022reducing} & 46.32 & 35.91 & - & - & - & 65.05 & 42.77 & - \\
        MATN~\cite{zhang2021multi} & 48.79 & 37.57 & - & - & - & - & - & - \\
        \textsubscript{This model doesn't have a name.}~\cite{panta2024cross} & 49.77 & 37.99 & - & - & - & - & - & - \\
        CDNet~\cite{ma2024disentangle} & 54.11 & 35.35 & 20.34 & 33.76 & 71.25 & 58.09 & 36.53 & - \\
        MS-DETR~\cite{jing-etal-2023-ms} & 47.66 & 37.36 & 25.81 & 35.09 & 68.68 & 57.72 &  37.40 & 50.12 \\
        CG-DETR~\cite{moon2023correlation} & 52.23 & 39.61 & 22.23 & 36.48 & 70.43 & 58.44 & 36.34 & 50.13 \\
        UVCOM~\cite{xiao2024bridging} & - & 36.39 & 23.32 & - & - & 59.25 & 36.64 & - \\
        R\textsuperscript{2}-Tuning~\cite{liu2024r} & 49.71 & 38.72 & 25.12 & 35.92 & 70.91 & 59.78 & 37.02 & 50.86 \\
        LLMEPET~\cite{jiang2024prior} & 52.73 & - & 22.78 & 36.55 & 70.91 & - & 36.49 & 50.25 \\
        UnLoc~\cite{yan2023unloc} & - & - & - & - & - & 60.80 & 38.40 & - \\
        MESM~\cite{liu2024towards} & 52.69 & 39.52 & - & 36.94 & - & 61.24 & 38.04 & - \\
        MCMN~\cite{han2023momentum} & 50.24 & 36.78 & - & - & - & \best{62.69} & \second{41.38} & - \\
        BAM-DETR~\cite{lee2023bam} & \second{56.69} & \second{41.54} & \best{26.77} & \second{39.31} & \second{72.93} & 59.95 & 39.38 & \second{52.33} \\
        
        % \rowcolor{my_blue}
        \textbf{LD-DETR (Ours)} & \best{57.61} & \best{44.31} & \second{26.24} & \best{40.30} & \best{73.92} & \second{62.58} & \best{41.56} & \best{53.44} \\
        
        \bottomrule[0.15em]
    \end{tabular}{}
    }
    \caption{\textbf{Video Moment Retrieval results on TACoS and Charades-STA. }The best result in each column is highlighted in \best{bold}, and the second best result in is highlighted in \second{underline}. $\dagger$ denotes training with audio modality. }
    \label{tab:charades_tacos_experiment}
\end{table*}
\endgroup

Table \ref{tab:tvsum_experiment} reports the LD-DETR's performance on Highlight Detection on TV-Sum dataset. The TV-Sum data set is too small, with only 4 training samples and 1 testing sample for each category. During training, the models quickly overfit. Even so, LD-DETR model still achieves not bad results on the TV-Sum dataset.

\begin{table*}[!t]
    \centering
    \scalebox{0.82}{
        \begin{tabular}{lccccccccccc}
        \toprule[0.15em]
        \textbf{Model} & VT & VU & GA & MS & PK & PR & FM & BK & BT & DS & \textbf{Avg.} \\

        \midrule[0.15em]
        
        sLSTM~\cite{zhang2016video} & 41.1 & 46.2 & 46.3 & 47.7 & 44.8 & 46.1 & 45.2 & 40.6 & 47.1 & 45.5 & 45.1 \\
        SG~\cite{yuan2019unsupervised} & 42.3 & 47.2 & 47.5 & 48.9 & 45.6 & 47.3 & 46.4 & 41.7 & 48.3 & 46.6 & 46.2 \\
        VESD~\cite{cai2018weakly} & 44.7 & 49.3 & 49.6 & 50.3 & 47.8 & 48.5 & 48.7 & 44.1 & 49.2 & 48.8 & 48.1 \\
        LIM-S~\cite{xiong2019less} & 55.9 & 42.9 & 61.2 & 54.0 & 60.3 & 47.5 & 43.2 & 66.3 & 69.1 & 62.6 & 56.3 \\
        Trailer~\cite{wang2020learning} & 61.3 & 54.6 & 65.7 & 60.8 & 59.1 & 70.1 & 58.2 & 64.7 & 65.6 & 68.1 & 62.8 \\
        MINI-Net$\dagger$~\cite{hong2020mini} & 80.6 & 68.3 & 78.2 & 81.8 & 78.1 & 65.8 & 75.8 & 75.0 & 80.2 & 65.5 & 73.2 \\
        SL-Module~\cite{xu2021cross} & 86.5 & 68.7 & 74.9 & 86.2 & 79.0 & 63.2 & 58.9 & 72.6 & 78.9 & 64.0 & 73.3 \\
        SA~\cite{badamdorj2021joint} & 83.4 & 64.7 & 84.4 & 86.5 & 70.3 & 67.5 & 66.9 & 68.1 & 95.0 & 60.8 & 74.8 \\
        SA\textsuperscript{+}$\dagger$~\cite{badamdorj2021joint} & 83.7 & 57.3 & 78.5 & 86.1 & 80.1 & 69.2 & 70.0 & 73.0 & 97.4 & 67.5 & 76.3 \\
        Joint-VA$\dagger$~\cite{badamdorj2021joint} & 83.7 & 57.3 & 78.5 & 86.1 & 80.1 & 69.2 & 70.0 & 73.0 & 97.4 & 67.5 & 76.3 \\
        TCG$\dagger$~\cite{ye2021temporal} & 85.0 & 71.4 & 81.9 & 78.6 & 80.2 & 75.5 & 71.6 & 77.3 & 78.6 & 68.1 & 76.8 \\
        PLD-VHD~\cite{wei2022learning} & 84.5 & 80.9 & 70.3 & 72.5 & 76.4 & 87.2 & 71.9 & 74.0 & 74.4 & 79.1 & 77.1 \\
        UniVTG~\cite{lin2023univtg} & 83.9 & 85.1 & 89.0 & 80.1 & 84.6 & 87.0 & 70.9 & 91.7 & 73.5 & 69.3 & 81.0 \\
        MH-DETR~\cite{xu2024mh} & 86.1 & 79.4 & 84.3 & 85.8 & 81.2 & 83.9 &  74.3 & 82.7 & 86.5 & 71.6 & 81.6 \\
        MIM$\dagger$~\cite{li2023mim} & 84.4 & 85.8 & 91.3 & 73.9 & 83.1 & 87.1 & 80.1 & 78.2 & 80.3 & 79.6 & 82.4 \\
        UMT$\dagger$~\cite{liu2022umt} & 87.5 & 81.5 & 88.2 & 78.8 & 81.4 & 87.0 & 76.0 & 86.9 & 84.4 & 79.6 & 83.1 \\
        VCSJT$\dagger$~\cite{zhou2024subtask} & 87.5 & 80.7 & 88.6 & 76.6 & 83.6 & 91.0 & 77.6 & 93.3 & 88.9 & 80.0 & 84.8 \\
        QD-DETR~\cite{moon2023query} & 88.2 & 87.4 & 85.6 & 85.0 & 85.8 & 86.9 & 76.4 & 91.3 & 89.2 & 73.7 & 85.0 \\
        
        % \rowcolor{my_blue}
        \textbf{LD-DETR (Ours)} & 83.1 & 90.3 & 91.5 & 82.5 & 87.7 & 83.5 & 79.0 & 90.4 & 86.1 & 77.3 & 85.1 \\
        
        UVCOM~\cite{xiao2024bridging} & 87.6 & 91.6 & 91.4 & 86.7 & 86.9 & 86.9 & 76.9 & 92.3 & 87.4 & 75.6 & 86.3 \\
        QD-DETR$\dagger$~\cite{moon2023query} & 87.6 & 91.7 & 90.2 & 88.3 & 84.1 & 88.3 & 78.7 & 91.2 & 87.8 & 77.7 & 86.6 \\
        CG-DETR~\cite{moon2023correlation} & 86.9 & 88.8 & 94.8 & 87.7 & 86.7 & 89.6 & 74.8 & 93.3 & 89.2 & 75.9 & 86.8 \\
        TR-DETR$\dagger$~\cite{sun2024tr} & 90.6 & 92.4 & 91.7 & 81.3 & 86.9 & 85.5 & 79.8 & 93.4 & 88.3 & 81.0 & 87.1 \\
        TaskWeave~\cite{yang2024task} & 88.2 & 90.8 & 93.3 & 87.5 & 87.0 & 82.0 & 80.9 & 92.9 & 89.5 & 81.2 & 87.3 \\
        LLMEPET~\cite{jiang2024prior} & 90.8 & 92.0 & 93.8 & 81.5 & 87.5 & 86.0 & 79.6 & 96.2 & 88.0 & 79.0 & 87.4 \\
        TR-DETR~\cite{sun2024tr} & 89.3 & 93.0 & 94.3 & 85.1 & 88.0 & 88.6 & 80.4 & 91.3 & 89.5 & 81.6 & 88.1 \\
        CDIM~\cite{1024497142.nh} & 85.5 & 95.8 & 90.3 & 90.0 & 88.4 & 88.1 & 79.2 & 97.1 & 88.0 & 80.5 & 88.3 \\
        \bottomrule[0.15em]
    \end{tabular}{}
    }
    \caption{\textbf{Highlight Detection results on TV-Sum. }
    % The best result in each column is highlighted in \best{bold}. 
    $\dagger$ denotes training with audio modality. }
    \label{tab:tvsum_experiment}
\end{table*}

\subsection{Ablation Studies}

\subsubsection{Ablation Studies on Distill Align}

Table \ref{tab:queue_align_experiment} shows the performance of Distill Align on multiple models. It demonstrates Distill Align as a plug-and-play method to improve the performance of multiple models. Through the Distill Align method, the video and text features mapped to the latent space are aligned to the same semantic space, and the overlapping semantic information in different training samples is taken into account during the alignment. 

\begingroup
\fontsize{9}{11}\selectfont
\begin{table*}[!h]
    \centering
    \scalebox{0.63}{
        \begin{tabular}{lccccccc}
        \toprule[0.15em]
        \multirow{3}{*}{\textbf{Setting}} & \multicolumn{5}{c}{\textbf{Video Moment Retrieval}} & \multicolumn{2}{c}{\textbf{Highlight Detection}} \\
        \cmidrule(l){2-6}  \cmidrule(l){7-8}
        & \multicolumn{2}{c}{R1} & \multicolumn{3}{c}{mAP} & \multicolumn{2}{c}{$>=$Very Good} \\
        \cmidrule(l){2-3} \cmidrule(l){4-6}  \cmidrule(l){7-8}
        & @0.5 & @0.7 & @0.5 & @0.75 & Avg. & mAP & HIT@1 \\

        \midrule[0.15em] 
        M-DETR~\cite{lei2021detecting} \source{NIPS'21} & 53.63$_{\pm\text{1.59}}$ & 35.78$_{\pm\text{1.25}}$ & 54.99$_{\pm\text{1.33}}$ & 31.28$_{\pm\text{1.07}}$ & 32.09$_{\pm\text{1.01}}$ & 36.41$_{\pm\text{0.40}}$ & 56.92$_{\pm\text{0.98}}$ \\
        M-DETR $+$ \textbf{Distill Align} & \best{55.56$_{\pm\text{1.01}}$} & \best{36.72$_{\pm\text{1.43}}$} & \best{56.42$_{\pm\text{0.76}}$} & \best{31.43$_{\pm\text{0.99}}$} & \best{32.66$_{\pm\text{0.73}}$} & \best{37.16$_{\pm\text{0.36}}$} & \best{58.62$_{\pm\text{1.19}}$} \\

        \midrule
        BM-DETR~\cite{jung2023overcoming} \source{arXiv'23} & 60.90$_{\pm\text{0.79}}$ & 44.10$_{\pm\text{1.43}}$ & 60.91$_{\pm\text{0.96}}$ & 39.03$_{\pm\text{1.46}}$ & \best{38.93$_{\pm\text{0.98}}$} & - & - \\
        BM-DETR $+$ \textbf{Distill Align} & \best{61.04$_{\pm\text{1.02}}$} & \best{44.23$_{\pm\text{1.04}}$} & \best{61.53$_{\pm\text{0.98}}$} & \best{39.12$_{\pm\text{1.45}}$} & 38.68$_{\pm\text{1.09}}$ & - & - \\

        \midrule
        QD-DETR~\cite{moon2023query} \source{CVPR'23} & 61.98$_{\pm\text{0.55}}$ & 47.30$_{\pm\text{0.69}}$ & 62.03$_{\pm\text{0.43}}$ & 41.96$_{\pm\text{0.66}}$ & 41.42$_{\pm\text{0.28}}$ & 38.92$_{\pm\text{0.30}}$ & 62.05$_{\pm\text{0.86}}$ \\
        QD-DETR $+$ \textbf{Distill Align} & \best{64.10$_{\pm\text{0.42}}$} & \best{48.53$_{\pm\text{0.70}}$} & \best{63.71$_{\pm\text{0.37}}$} & \best{43.55$_{\pm\text{0.59}}$} & \best{42.80$_{\pm\text{0.46}}$} & \best{39.58$_{\pm\text{0.40}}$} & \best{63.21$_{\pm\text{1.05}}$} \\

        \midrule
        CG-DETR~\cite{moon2023correlation} \source{arXiv'23} & 65.92$_{\pm\text{0.22}}$ & 50.44$_{\pm\text{0.54}}$ & 65.50$_{\pm\text{0.22}}$ & 45.62$_{\pm\text{0.68}}$ & 44.76$_{\pm\text{0.26}}$ & 40.34$_{\pm\text{0.20}}$ & 65.12$_{\pm\text{0.64}}$ \\
        CG-DETR $+$ \textbf{Distill Align} & \best{66.11$_{\pm\text{0.79}}$} & \best{51.18$_{\pm\text{0.89}}$} & \best{65.65$_{\pm\text{0.61}}$} & \best{46.12$_{\pm\text{0.74}}$} & \best{45.23$_{\pm\text{0.61}}$} & \best{40.50$_{\pm\text{0.23}}$} & \best{65.92$_{\pm\text{1.04}}$} \\

        \midrule
        TR-DETR~\cite{sun2024tr} \source{AAAI'24} & 66.56$_{\pm\text{1.06}}$ & 50.13$_{\pm\text{0.89}}$ & 65.70$_{\pm\text{0.79}}$ & 45.10$_{\pm\text{0.78}}$ & 44.33$_{\pm\text{0.51}}$ & 40.88$_{\pm\text{0.19}}$ & 65.54$_{\pm\text{0.45}}$ \\
        TR-DETR $-$ \textsc{LGMA} & 62.28$_{\pm\text{1.08}}$ & 46.99$_{\pm\text{1.21}}$ & 62.38$_{\pm\text{0.72}}$ & 62.16$_{\pm\text{1.41}}$ & 41.56$_{\pm\text{1.12}}$ & 39.16$_{\pm\text{0.32}}$ & 62.13$_{\pm\text{0.46}}$ \\
        TR-DETR $+$ \textbf{Distill Align} & 66.31$_{\pm\text{0.60}}$ & 49.92$_{\pm\text{0.73}}$ & 65.33$_{\pm\text{0.74}}$ & 44.14$_{\pm\text{0.81}}$ & 43.60$_{\pm\text{0.94}}$ & \best{40.89$_{\pm\text{0.26}}$} & \best{65.79$_{\pm\text{1.02}}$} \\
        TR-DETR $-$ \textsc{LGMA} $+$ \textbf{Distill Align} & \best{67.14$_{\pm\text{0.43}}$} & \best{51.17$_{\pm\text{0.37}}$} & \best{66.21$_{\pm\text{0.24}}$} & \best{45.57$_{\pm\text{0.42}}$} & \best{44.89$_{\pm\text{0.32}}$} & 40.77$_{\pm\text{0.31}}$ & 65.33$_{\pm\text{0.71}}$ \\

        \midrule
        UVCOM~\cite{xiao2024bridging} \source{CVPR'24} & 64.36$_{\pm\text{0.44}}$ & 50.21$_{\pm\text{1.06}}$ & 63.99$_{\pm\text{0.27}}$ & 45.52$_{\pm\text{0.68}}$ & 44.77$_{\pm\text{0.53}}$ & 39.85$_{\pm\text{0.21}}$ & 63.82$_{\pm\text{1.05}}$ \\
        UVCOM $+$ \textbf{Distill Align} & \best{66.39$_{\pm\text{0.37}}$} & \best{51.54$_{\pm\text{0.22}}$} & \best{65.24$_{\pm\text{0.49}}$} & \best{46.18$_{\pm\text{0.38}}$} & \best{45.39$_{\pm\text{0.29}}$} & \best{40.72$_{\pm\text{0.14}}$} & \best{65.20$_{\pm\text{0.73}}$} \\
        
        \bottomrule[0.15em]
    \end{tabular}{}
    }
    \caption{\textbf{Distill Align as a plug-and-play method can help multiple models achieve better results. }\textsc{LGMA} is the abbreviation of Local-Global Multi-Modal Alignment which is a method used in the TR-DETR~\cite{sun2024tr} model. The best result on each baseline in each column is highlighted in \best{bold}. }
    \label{tab:queue_align_experiment}
\end{table*}
\endgroup

\begin{figure}[!h]
\centering
\includegraphics[width=0.45\textwidth]{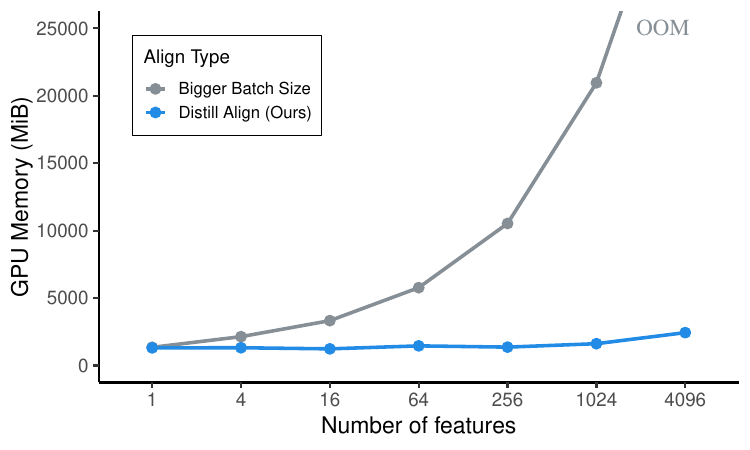}
\includegraphics[width=0.45\textwidth]{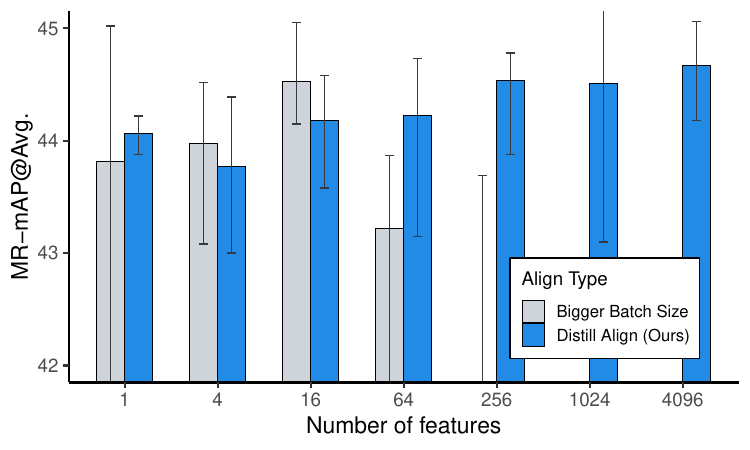}
\caption{\textbf{The Distill Align method can introduce more samples for comparative learning without taking up too much GPU memory. A}s the queue length increases, the model results get better. We visualized the GPU memory usage and the results on the QVHighlights dataset of the two methods when the number of features involved in contrastive learning increases. The experiment is conducted on an NVIDIA GeForce RTX 4090 with 24,564MiB GPU memory using TR-DETR~\cite{sun2024tr} as the baseline. The x-axis represents the relative number of features involved in contrastive learning, where x-axis = 1 represents a batch size of 32 and a queue length of 0. “OOM” means Out of Memory. This figure shows the same set of experiments as Table \ref{tab:queue_align_experiment2}. }
\label{fig:queue_align_experiment2}
\end{figure}

Figure \ref{fig:queue_align_experiment2} and Table \ref{tab:queue_align_experiment2} show that the Distill Align method enables the model to achieve better results without occupying too much GPU memory. It can be seen that, when use Bigger Batch Size, as the batch size increases, the effect of the model becomes better. However, bigger batch size causes the GPU memory occupied by the model to increase, and it is more difficult for the model to converge as the batch size increases. Specifically, when batch size = 1,024, we conducted 5 experiments, and only one of them successfully converged the model. When using Distill Align, as the queue length increases, the effect of the model continues to get better, while the GPU memory occupied is much smaller.

\begingroup
\fontsize{9}{11}\selectfont
\begin{table*}[!h]
    \centering
    \scalebox{0.58}{
        \begin{tabular}{ccccccccc}
        \toprule[0.15em]
        \multirow{3}{*}{\textbf{Batch Size / Queue Length}} & \multirow{3}{*}{\textbf{GPU Memory}} & \multicolumn{5}{c}{\textbf{Video Moment Retrieval}} & \multicolumn{2}{c}{\textbf{Highlight Detection}} \\
        \cmidrule(l){3-7}  \cmidrule(l){8-9}
        & & \multicolumn{2}{c}{R1} & \multicolumn{3}{c}{mAP} & \multicolumn{2}{c}{$>=$Very Good} \\
        \cmidrule(l){3-4} \cmidrule(l){5-7}  \cmidrule(l){8-9}
        & & @0.5 & @0.7 & @0.5 & @0.75 & Avg. & mAP & HIT@1 \\

        \midrule[0.15em]
        \multicolumn{9}{l}{\textbf{Bigger Batch Size}} \\
        \midrule

        bs = 32, ql = 0 & 1,350 & 66.19$_{\pm\text{1.41}}$ & 49.74$_{\pm\text{1.70}}$ & 65.27$_{\pm\text{1.11}}$ & 44.12$_{\pm\text{1.41}}$ & 43.81$_{\pm\text{1.19}}$ & 40.72$_{\pm\text{0.17}}$ & 65.19$_{\pm\text{1.23}}$ \\
        bs = 64, ql = 0 & 2,140 & 66.89$_{\pm\text{0.75}}$ & 50.07$_{\pm\text{0.82}}$ & 65.73$_{\pm\text{0.39}}$ & 44.53$_{\pm\text{0.54}}$ & 43.98$_{\pm\text{0.50}}$ & 40.99$_{\pm\text{0.29}}$ & 65.86$_{\pm\text{0.84}}$ \\
        bs = 128, ql = 0 & 3,328 & 66.81$_{\pm\text{0.70}}$ & 50.93$_{\pm\text{0.46}}$ & 65.93$_{\pm\text{0.46}}$ & 45.00$_{\pm\text{0.43}}$ & 44.53$_{\pm\text{0.33}}$ & 40.92$_{\pm\text{0.20}}$ & 65.77$_{\pm\text{1.26}}$ \\
        bs = 256, ql = 0 & 5,772 & 66.62$_{\pm\text{0.59}}$ & 49.56$_{\pm\text{0.82}}$ & 65.27$_{\pm\text{0.51}}$ & 43.46$_{\pm\text{0.66}}$ & 43.22$_{\pm\text{0.83}}$ & 40.76$_{\pm\text{0.16}}$ & 65.38$_{\pm\text{0.52}}$ \\
        bs = 512, ql = 0 & 10,530 & 65.20$_{\pm\text{1.92}}$ & 47.38$_{\pm\text{2.31}}$ & 63.49$_{\pm\text{1.85}}$ & 41.35$_{\pm\text{2.33}}$ & 40.90$_{\pm\text{2.00}}$ & 40.58$_{\pm\text{0.17}}$ & 64.94$_{\pm\text{1.16}}$ \\
        bs = 1,024, ql = 0 & 20,956 & 19.15$_{\pm\text{13.66}}$ & 8.70$_{\pm\text{8.05}}$ & 26.29$_{\pm\text{10.96}}$ & 9.45$_{\pm\text{6.48}}$ & 11.73$_{\pm\text{6.34}}$ & 27.13$_{\pm\text{5.20}}$ & 37.18$_{\pm\text{10.86}}$ \\
        bs = 2,048, ql = 0 & Out of Memory & - & - & - & - & - & - & - \\

        \midrule
        \multicolumn{9}{l}{\textbf{Distill Align (Ours)}} \\
        \midrule
        
        bs = 32, ql = 0 & 1,312 & 66.53$_{\pm\text{0.96}}$ & 49.90$_{\pm\text{0.31}}$ & 65.70$_{\pm\text{0.89}}$ & 44.57$_{\pm\text{0.48}}$ & 44.07$_{\pm\text{0.13}}$ & 40.52$_{\pm\text{0.24}}$ & 64.73$_{\pm\text{0.95}}$ \\
        bs = 32, ql = 96 & 1,318 & 66.53$_{\pm\text{0.32}}$ & 49.34$_{\pm\text{0.33}}$ & 65.61$_{\pm\text{0.85}}$ & 44.13$_{\pm\text{0.35}}$ & 43.77$_{\pm\text{0.49}}$ & 40.50$_{\pm\text{0.16}}$ & 64.66$_{\pm\text{0.24}}$ \\
        bs = 32, ql = 480 & 1,236 & 67.11$_{\pm\text{1.49}}$ & 50.48$_{\pm\text{1.03}}$ & \best{66.28$_{\pm\text{1.03}}$} & 45.10$_{\pm\text{0.51}}$ & 44.18$_{\pm\text{0.41}}$ & 40.92$_{\pm\text{0.23}}$ & 65.91$_{\pm\text{0.97}}$ \\
        bs = 32, ql = 2,016 & 1,458 & 66.76$_{\pm\text{0.57}}$ & 50.48$_{\pm\text{0.40}}$ & 65.73$_{\pm\text{0.47}}$ & 44.83$_{\pm\text{0.68}}$ & 44.22$_{\pm\text{0.57}}$ & 40.81$_{\pm\text{0.39}}$ & 65.09$_{\pm\text{0.70}}$ \\
        bs = 32, ql = 8,160 & 1,364 & 66.74$_{\pm\text{0.84}}$ & \best{50.81$_{\pm\text{0.50}}$} & 66.07$_{\pm\text{0.42}}$ & \best{45.30$_{\pm\text{0.52}}$} & 44.53$_{\pm\text{0.34}}$ & 40.76$_{\pm\text{0.19}}$ & \best{65.95$_{\pm\text{0.42}}$} \\
        bs = 32, ql = 32,736 & 1,620 & 67.00$_{\pm\text{0.75}}$ & 50.71$_{\pm\text{1.10}}$ & 66.10$_{\pm\text{0.67}}$ & 45.06$_{\pm\text{0.99}}$ & 44.51$_{\pm\text{0.80}}$ & \best{41.00$_{\pm\text{0.21}}$} & 65.64$_{\pm\text{0.53}}$ \\
        bs = 32, ql = 131,040 & 2,440 & \best{67.14$_{\pm\text{0.59}}$} & 50.63$_{\pm\text{0.48}}$ & 65.91$_{\pm\text{0.32}}$ & 45.21$_{\pm\text{0.60}}$ & \best{44.67$_{\pm\text{0.30}}$} & \best{41.00$_{\pm\text{0.23}}$} & 65.86$_{\pm\text{0.82}}$ \\
        
        \bottomrule[0.15em]
    \end{tabular}{}
    }
    \caption{\textbf{Compared with using a Bigger Batch Size, using Distill Align saves GPU memory and will get better results. }This table shows the GPU memory usage and training results when using different batch sizes or queue lengths, using a Bigger Batch Size and using Distill Align. “bs” means batch size, “ql” means queue length. The experiment is conducted on an NVIDIA GeForce RTX 4090 with 24,564MiB GPU memory using TR-DETR~\cite{sun2024tr} as the baseline. The best result in each column is highlighted in \best{bold}. This table shows the same set of experiments as Figure \ref{fig:queue_align_experiment2}. }
    \label{tab:queue_align_experiment2}
\end{table*}
\endgroup

\begin{figure}[!h]
\centering
\includegraphics[width=0.6\textwidth]{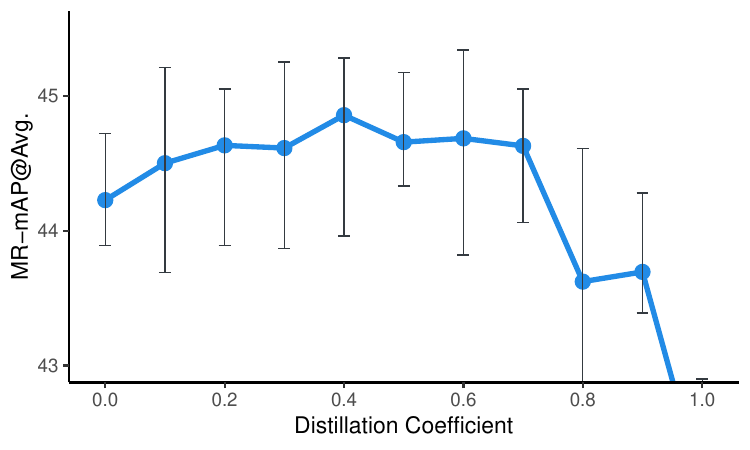}
\caption{\textbf{This figure shows the impact of distillation coefficient on multimodal alignment in Distill Align. }When Distillation Coefficient = 0.0, it is equivalent to not enabling distillation. When a certain degree of distillation is used, the performance of the model becomes better. This figure shows the same set of experiments as Table \ref{tab:queue_align_alpha_experiment}. }
\label{fig:queue_align_alpha_experiment}
\end{figure}

Figure \ref{fig:queue_align_alpha_experiment} and Table \ref{tab:queue_align_alpha_experiment} show the impact of distillation on multimodal alignment in Distill Align. Among them, when Distillation Coefficient = 0.0, it is equivalent to not enabling distillation. It can be seen that when a certain degree of distillation is used, the overlapping semantic information between different training samples is taken into account by multi-modal alignment, and the performance of the model becomes better. 

\begingroup
\fontsize{9}{11}\selectfont
\begin{table*}[!h]
    \centering
    \scalebox{0.70}{
        \begin{tabular}{cccccccc}
        \toprule[0.15em]
        \multirow{3}{*}{\textbf{Distillation Coefficient}} & \multicolumn{5}{c}{\textbf{Video Moment Retrieval}} & \multicolumn{2}{c}{\textbf{Highlight Detection}} \\
        \cmidrule(l){2-6}  \cmidrule(l){7-8}
        & \multicolumn{2}{c}{R1} & \multicolumn{3}{c}{mAP} & \multicolumn{2}{c}{$>=$Very Good} \\
        \cmidrule(l){2-3} \cmidrule(l){4-6}  \cmidrule(l){7-8}
        & @0.5 & @0.7 & @0.5 & @0.75 & Avg. & mAP & HIT@1 \\

        \midrule[0.15em] 
        0.0 & 66.28$_{\pm\text{1.38}}$ & 49.91$_{\pm\text{0.44}}$ & 65.64$_{\pm\text{0.74}}$ & 44.60$_{\pm\text{0.68}}$ & 44.23$_{\pm\text{0.27}}$ & 40.81$_{\pm\text{0.21}}$ & 65.03$_{\pm\text{0.78}}$ \\
        0.1 & 67.20$_{\pm\text{0.64}}$ & 50.50$_{\pm\text{0.78}}$ & 66.20$_{\pm\text{0.40}}$ & 44.83$_{\pm\text{0.64}}$ & 44.50$_{\pm\text{0.60}}$ & 40.85$_{\pm\text{0.20}}$ & 65.58$_{\pm\text{0.89}}$ \\
        0.2 & 67.15$_{\pm\text{0.99}}$ & 50.50$_{\pm\text{0.26}}$ & 66.28$_{\pm\text{0.61}}$ & 45.24$_{\pm\text{0.39}}$ & 44.63$_{\pm\text{0.40}}$ & 40.98$_{\pm\text{0.18}}$ & 66.00$_{\pm\text{0.65}}$ \\
        0.3 & 67.07$_{\pm\text{0.59}}$ & 50.62$_{\pm\text{0.42}}$ & 66.19$_{\pm\text{0.54}}$ & 45.29$_{\pm\text{0.52}}$ & 44.61$_{\pm\text{0.49}}$ & \best{41.14$_{\pm\text{0.16}}$} & 65.78$_{\pm\text{1.26}}$ \\
        0.4 & \best{67.59$_{\pm\text{0.71}}$} & \best{50.89$_{\pm\text{0.82}}$} & \best{66.48$_{\pm\text{0.37}}$} & \best{45.50$_{\pm\text{0.58}}$} & \best{44.86$_{\pm\text{0.50}}$} & 41.09$_{\pm\text{0.22}}$ & \best{66.32$_{\pm\text{0.47}}$} \\
        0.5 & 66.76$_{\pm\text{0.89}}$ & 50.46$_{\pm\text{0.62}}$ & 66.16$_{\pm\text{0.54}}$ & 45.06$_{\pm\text{0.50}}$ & 44.66$_{\pm\text{0.28}}$ & 40.83$_{\pm\text{0.37}}$ & 65.56$_{\pm\text{0.62}}$ \\
        0.6 & 66.61$_{\pm\text{0.43}}$ & 50.55$_{\pm\text{0.59}}$ & 65.89$_{\pm\text{0.54}}$ & 45.32$_{\pm\text{0.56}}$ & 44.68$_{\pm\text{0.49}}$ & 40.65$_{\pm\text{0.20}}$ & 64.55$_{\pm\text{0.67}}$ \\
        0.7 & 66.59$_{\pm\text{0.28}}$ & 50.52$_{\pm\text{0.81}}$ & 65.71$_{\pm\text{0.22}}$ & 45.14$_{\pm\text{0.49}}$ & 44.63$_{\pm\text{0.32}}$ & 41.05$_{\pm\text{0.19}}$ & 65.43$_{\pm\text{0.42}}$ \\
        0.8 & 65.72$_{\pm\text{0.94}}$ & 49.55$_{\pm\text{1.09}}$ & 64.87$_{\pm\text{0.57}}$ & 44.05$_{\pm\text{1.51}}$ & 43.62$_{\pm\text{1.09}}$ & 40.60$_{\pm\text{0.13}}$ & 64.80$_{\pm\text{0.50}}$ \\
        0.9 & 65.25$_{\pm\text{0.70}}$ & 49.47$_{\pm\text{0.33}}$ & 64.75$_{\pm\text{0.22}}$ & 44.17$_{\pm\text{0.36}}$ & 43.69$_{\pm\text{0.33}}$ & 40.23$_{\pm\text{0.34}}$ & 63.88$_{\pm\text{1.25}}$ \\
        1.0 & 62.83$_{\pm\text{0.82}}$ & 47.26$_{\pm\text{0.85}}$ & 63.08$_{\pm\text{0.49}}$ & 42.54$_{\pm\text{0.82}}$ & 42.09$_{\pm\text{0.58}}$ & 39.37$_{\pm\text{0.21}}$ & 61.87$_{\pm\text{0.39}}$ \\
        % 0.0 & $_{\pm\text{}}$ & $_{\pm\text{}}$ & $_{\pm\text{}}$ & $_{\pm\text{}}$ & $_{\pm\text{}}$ & $_{\pm\text{}}$ & $_{\pm\text{}}$ \\
        
        \bottomrule[0.15em]
    \end{tabular}{}
    }
    \caption{\textbf{This table shows the impact of distillation on multimodal alignment in Distill Align. }The best result on each baseline in each column is highlighted in \best{bold}. When Distillation Coefficient = 0.0, it is equivalent to not enabling distillation. When a certain degree of distillation is used, the performance of the model becomes better. This table shows the same set of experiments as Figure \ref{fig:queue_align_alpha_experiment}. }
    \label{tab:queue_align_alpha_experiment}
\end{table*}
\endgroup

\subsubsection{Ablation Studies on Convolutional Fuser} 

Table \ref{tab:1dconvexperiment1} shows the impact of the number of convolutional layers in Convolutional Fuser. The motion information in the video is often contained in several local clips. Convolutional layers can effectively capture the local information in the video. Compared with UVCOM~\cite{xiao2024bridging} (a-f), our method (g-m) can better utilize the ability of convolutional layers to extract local features. In UVCOM, as the number of convolutional layers increases, the performance of the model does not get better. But in our method, as the number of convolutional layers increases, the performance of the model becomes better. Compared with not using convolutional layers (g), just adding one convolutional layer (h) can greatly improve the model effect. As the number of convolutional layers added increases, the model is able to capture local information in the video more effectively and the effect gradually improves until it reaches the peak (l).

\begingroup
\fontsize{9}{11}\selectfont
\begin{table*}[!h]
    \centering
    \scalebox{0.74}{
        \begin{tabular}{ccccccccc}
        \toprule[0.15em]
        \multirow{3}{*}{\textbf{Setting}} & \multirow{3}{*}{\# Conv} & \multicolumn{5}{c}{\textbf{Video Moment Retrieval}} & \multicolumn{2}{c}{\textbf{Highlight Detection}} \\
        \cmidrule(l){3-7}  \cmidrule(l){8-9}
        & & \multicolumn{2}{c}{R1} & \multicolumn{3}{c}{mAP} & \multicolumn{2}{c}{$>=$Very Good} \\
        \cmidrule(l){3-4} \cmidrule(l){5-7}  \cmidrule(l){8-9}
        & & @0.5 & @0.7 & @0.5 & @0.75 & Avg. & mAP & HIT@1 \\

        \midrule[0.15em]
        \multicolumn{9}{l}{\textbf{UVCOM}~\cite{xiao2024bridging} \source{CVPR'24}} \\
        \midrule
        
        (a) & 0 & 63.68$_{\pm\text{0.43}}$ & 49.29$_{\pm\text{0.86}}$ & 63.50$_{\pm\text{0.46}}$ & 43.81$_{\pm\text{0.44}}$ & 43.40$_{\pm\text{0.41}}$ & 39.40$_{\pm\text{0.13}}$ & 63.22$_{\pm\text{0.46}}$ \\
        (b) & 1 & 64.36$_{\pm\text{0.44}}$ & 50.21$_{\pm\text{1.06}}$ & 63.99$_{\pm\text{0.27}}$ & 45.52$_{\pm\text{0.68}}$ & 44.72$_{\pm\text{0.53}}$ & 39.85$_{\pm\text{0.21}}$ & 63.82$_{\pm\text{1.05}}$ \\
        (c) & 2 & 61.68$_{\pm\text{0.59}}$ & 47.51$_{\pm\text{0.54}}$ & 61.72$_{\pm\text{0.74}}$ & 42.66$_{\pm\text{0.53}}$ & 42.40$_{\pm\text{0.47}}$ & 38.64$_{\pm\text{0.43}}$ & 60.90$_{\pm\text{0.81}}$ \\
        (d) & 4 & 61.89$_{\pm\text{0.63}}$ & 47.54$_{\pm\text{0.85}}$ & 61.63$_{\pm\text{0.82}}$ & 42.46$_{\pm\text{0.78}}$ & 42.46$_{\pm\text{0.81}}$ & 38.52$_{\pm\text{0.13}}$ & 60.54$_{\pm\text{0.75}}$ \\
        % () & 6 & 60.66$_{\pm\text{0.36}}$ & 46.66$_{\pm\text{0.61}}$ & 61.02$_{\pm\text{0.41}}$ & 42.02$_{\pm\text{1.07}}$ & 41.87$_{\pm\text{0.79}}$ & 38.25$_{\pm\text{0.27}}$ & 59.47$_{\pm\text{0.82}}$ \\
        (e) & 8 & 60.87$_{\pm\text{0.23}}$ & 46.95$_{\pm\text{0.63}}$ & 61.41$_{\pm\text{0.32}}$ & 42.33$_{\pm\text{0.50}}$ & 42.28$_{\pm\text{0.23}}$ & 38.39$_{\pm\text{0.23}}$ & 60.22$_{\pm\text{0.92}}$ \\
        % () & 10 & $_{\pm\text{}}$ & $_{\pm\text{}}$ & $_{\pm\text{}}$ & $_{\pm\text{}}$ & $_{\pm\text{}}$ & $_{\pm\text{}}$ & $_{\pm\text{}}$ \\
        (f) & 16 & 61.46$_{\pm\text{0.60}}$ & 47.13$_{\pm\text{0.42}}$ & 61.76$_{\pm\text{0.25}}$ & 43.07$_{\pm\text{0.62}}$ & 42.57$_{\pm\text{0.33}}$ & 38.19$_{\pm\text{0.23}}$ & 59.80$_{\pm\text{0.77}}$ \\

        \midrule
        \multicolumn{9}{l}{\textbf{LD-DETR (Ours)}} \\
        \midrule
        
        (g) & 0 & 67.07$_{\pm\text{1.54}}$ & 50.96$_{\pm\text{2.17}}$ & 65.79$_{\pm\text{1.72}}$ & 45.17$_{\pm\text{2.10}}$ & 44.18$_{\pm\text{1.91}}$ & 41.42$_{\pm\text{0.09}}$ & 66.30$_{\pm\text{0.29}}$ \\
        (h) & 1 & 68.14$_{\pm\text{0.26}}$ & 52.01$_{\pm\text{1.28}}$ & 66.70$_{\pm\text{0.38}}$ & 46.24$_{\pm\text{1.18}}$ & 45.71$_{\pm\text{0.96}}$ & 41.70$_{\pm\text{0.20}}$ & \best{67.41$_{\pm\text{0.95}}$} \\
        (i) & 2 & 68.10$_{\pm\text{0.63}}$ & 52.04$_{\pm\text{0.45}}$ & 67.49$_{\pm\text{0.78}}$ & 47.01$_{\pm\text{0.25}}$ & 46.73$_{\pm\text{0.32}}$ & 41.65$_{\pm\text{0.24}}$ & 66.62$_{\pm\text{0.62}}$ \\
        (j) & 4 & 68.77$_{\pm\text{0.40}}$ & 52.88$_{\pm\text{0.66}}$ & 68.00$_{\pm\text{0.38}}$ & 47.55$_{\pm\text{0.66}}$ & 47.21$_{\pm\text{0.46}}$ & \best{41.83$_{\pm\text{0.15}}$} & 67.30$_{\pm\text{1.33}}$ \\
        (k) & 8 & 68.52$_{\pm\text{0.90}}$ & 52.22$_{\pm\text{1.06}}$ & 67.70$_{\pm\text{0.62}}$ & 67.14$_{\pm\text{0.90}}$ & 47.28$_{\pm\text{0.66}}$ & 41.55$_{\pm\text{0.24}}$ & 66.41$_{\pm\text{0.61}}$ \\
        (l) & 10 & \best{69.01$_{\pm\text{1.09}}$} & \best{53.19$_{\pm\text{0.38}}$} & \best{68.43$_{\pm\text{0.83}}$} & \best{48.25$_{\pm\text{0.59}}$} & \best{47.93$_{\pm\text{0.39}}$} & 41.66$_{\pm\text{0.15}}$ & 66.80$_{\pm\text{0.96}}$ \\
        (m) & 16 & 68.63$_{\pm\text{0.97}}$ & 52.89$_{\pm\text{0.62}}$ & 68.10$_{\pm\text{0.64}}$ & 47.56$_{\pm\text{0.36}}$ & 47.24$_{\pm\text{0.46}}$ & 41.79$_{\pm\text{0.15}}$ & 67.33$_{\pm\text{0.42}}$ \\
        % (a) & 0 & $_{\pm\text{}}$ & $_{\pm\text{}}$ & $_{\pm\text{}}$ & $_{\pm\text{}}$ & $_{\pm\text{}}$ & $_{\pm\text{}}$ & $_{\pm\text{}}$ \\
        
        \bottomrule[0.15em]
    \end{tabular}{}
    }
    \caption{\textbf{This table shows the impact of the number of convolutional layers used on model performance. }“\# Conv” means the number of convolutional layers. The best result in each column is highlighted in \best{bold}. It can be noticed that on LD-DETR, with the help of Convolutional Fuser, just adding one convolutional layer can significantly improve the model performance. As the number of convolutional layers increases, the performance of the model gets better until it reaches a limit. On UVCOM~\cite{xiao2024bridging}, however, adding convolutional layers does not help the model to perform better. }
    \label{tab:1dconvexperiment1}
\end{table*}
\endgroup

Table \ref{tab:1dconvexperiment2} and Table \ref{tab:1dconvexperiment3} show ablation experiments on Convolutional Fuser. Whether you  the order of the methods or remove any of them, the performance of the model will degrade.

\begingroup
\fontsize{9}{11}\selectfont
\begin{table*}[!h]
    \centering
    \scalebox{0.62}{
        \begin{tabular}{ccccccccc}
        \toprule[0.15em]
        \multirow{3}{*}{\textbf{Setting}} & \multirow{3}{*}{\textbf{Position}} & \multicolumn{5}{c}{\textbf{Video Moment Retrieval}} & \multicolumn{2}{c}{\textbf{Highlight Detection}} \\
        \cmidrule(l){3-7}  \cmidrule(l){8-9}
        & & \multicolumn{2}{c}{R1} & \multicolumn{3}{c}{mAP} & \multicolumn{2}{c}{$>=$Very Good} \\
        \cmidrule(l){3-4} \cmidrule(l){5-7}  \cmidrule(l){8-9}
        & & @0.5 & @0.7 & @0.5 & @0.75 & Avg. & mAP & HIT@1 \\

        \midrule[0.15em]
        
        (a) & Before \textsc{V2TEx} & 58.72$_{\pm\text{2.48}}$ & 42.43$_{\pm\text{3.08}}$ & 58.45$_{\pm\text{2.87}}$ & 37.44$_{\pm\text{2.88}}$ & 37.14$_{\pm\text{2.84}}$ & 38.20$_{\pm\text{0.62}}$ & 60.39$_{\pm\text{1.60}}$ \\
        (b) & Between \textsc{V2TEx} \& \textsc{T2VEn} & 66.64$_{\pm\text{1.21}}$ & 49.61$_{\pm\text{1.43}}$ & 65.84$_{\pm\text{1.00}}$ & 43.99$_{\pm\text{1.19}}$ & 43.51$_{\pm\text{1.15}}$ & 41.20$_{\pm\text{0.19}}$ & 66.36$_{\pm\text{0.48}}$ \\
        (c) & Between \textsc{T2VEn} \& \textsc{TrEn1} & 68.87$_{\pm\text{0.68}}$ & 52.96$_{\pm\text{0.32}}$ & 68.07$_{\pm\text{0.56}}$ & 47.54$_{\pm\text{0.41}}$ & 47.51$_{\pm\text{0.39}}$ & \best{41.81$_{\pm\text{0.13}}$} & 66.79$_{\pm\text{0.50}}$ \\
        (d) & Between \textsc{TrEn1} \& \textsc{TrEn2} & \best{69.01$_{\pm\text{1.09}}$} & \best{53.19$_{\pm\text{0.38}}$} & \best{68.43$_{\pm\text{0.83}}$} & \best{48.25$_{\pm\text{0.59}}$} & \best{47.93$_{\pm\text{0.39}}$} & 41.66$_{\pm\text{0.15}}$ & \best{66.80$_{\pm\text{0.96}}$} \\
        (e) & After \textsc{TrEn2} & 67.62$_{\pm\text{0.96}}$ & 52.23$_{\pm\text{0.53}}$ & 67.17$_{\pm\text{0.17}}$ & 46.85$_{\pm\text{0.23}}$ & 46.57$_{\pm\text{0.19}}$ & 41.56$_{\pm\text{0.28}}$ & 66.26$_{\pm\text{0.52}}$ \\
        
        \bottomrule[0.15em]
    \end{tabular}{}
    }
    \caption{\textbf{This table shows the performances of the model when Convolutional Blocks are placed in different locations in Convolutional Fuser. }It can be noticed that the model performs best when Convolutional Blocks are placed between Transformer Encoder 1 and Transformer Encoder 2. In this table, “\textsc{V2TEx}” means V2T Extractor, “\textsc{T2VEn}” means T2V Encoder, “\textsc{TrEn1}” means Transformer Encoder 1, “\textsc{TrEn2}” means Transformer Encoder 2. The best result in each column is highlighted in \best{bold}. }
    \label{tab:1dconvexperiment2}
\end{table*}
\endgroup

\begingroup
\fontsize{9}{11}\selectfont
\begin{table*}[!h]
    \centering
    \scalebox{0.53}{
        \begin{tabular}{ccccccccccccc}
        \toprule[0.15em]
        \multirow{3}{*}{\textbf{Setting}} & \multirow{3}{*}{\textsc{V2TEx}} & \multirow{3}{*}{\textsc{T2VEn}} & \multirow{3}{*}{\textsc{TrEn1}} & \multirow{3}{*}{\textsc{ConBl}} & \multirow{3}{*}{\textsc{TrEn2}} & \multicolumn{5}{c}{\textbf{Video Moment Retrieval}} & \multicolumn{2}{c}{\textbf{Highlight Detection}} \\
        \cmidrule(l){7-11}  \cmidrule(l){12-13}
        & & & & & & \multicolumn{2}{c}{R1} & \multicolumn{3}{c}{mAP} & \multicolumn{2}{c}{$>=$Very Good} \\
        \cmidrule(l){7-8} \cmidrule(l){9-11}  \cmidrule(l){12-13}
        & & & & & & @0.5 & @0.7 & @0.5 & @0.75 & Avg. & mAP & HIT@1 \\

        \midrule[0.15em]
        
        (a) & & \checkmark & \checkmark & \checkmark & \checkmark & 63.82$_{\pm\text{0.66}}$ & 49.24$_{\pm\text{0.77}}$ & 63.94$_{\pm\text{0.11}}$ & 44.65$_{\pm\text{0.28}}$ & 44.28$_{\pm\text{0.26}}$ & 39.95$_{\pm\text{0.27}}$ & 62.92$_{\pm\text{1.02}}$ \\
        (b) & \checkmark & & \checkmark & \checkmark & \checkmark & 68.17$_{\pm\text{0.88}}$ & 52.41$_{\pm\text{0.96}}$ & 67.11$_{\pm\text{0.73}}$ & 46.74$_{\pm\text{0.96}}$ & 46.53$_{\pm\text{0.50}}$ & 41.50$_{\pm\text{0.25}}$ & 66.27$_{\pm\text{0.95}}$ \\
        (c) & \checkmark & \checkmark & & \checkmark & \checkmark & 68.09$_{\pm\text{0.65}}$ & 52.61$_{\pm\text{0.84}}$ & 67.53$_{\pm\text{0.49}}$ & 47.37$_{\pm\text{0.42}}$ & 47.03$_{\pm\text{0.17}}$ & 41.43$_{\pm\text{0.15}}$ & 66.62$_{\pm\text{0.43}}$ \\
        (d) & \checkmark & \checkmark & \checkmark & & \checkmark & 67.07$_{\pm\text{1.54}}$ & 50.96$_{\pm\text{2.17}}$ & 65.79$_{\pm\text{1.72}}$ & 45.17$_{\pm\text{2.10}}$ & 44.18$_{\pm\text{1.91}}$ & 41.42$_{\pm\text{0.09}}$ & 66.30$_{\pm\text{0.29}}$ \\
        (e) & \checkmark & \checkmark & \checkmark & \checkmark & & 68.36$_{\pm\text{0.31}}$ & 52.10$_{\pm\text{0.36}}$ & 67.67$_{\pm\text{0.17}}$ & 47.33$_{\pm\text{0.65}}$ & 47.07$_{\pm\text{0.36}}$ & 41.30$_{\pm\text{0.24}}$ & 66.75$_{\pm\text{0.92}}$ \\
        
        % \midrule
        
        (f) & \checkmark & \checkmark & \checkmark & \checkmark & \checkmark & \best{69.01$_{\pm\text{1.09}}$} & \best{53.19$_{\pm\text{0.38}}$} & \best{68.43$_{\pm\text{0.83}}$} & \best{48.25$_{\pm\text{0.59}}$} & \best{47.93$_{\pm\text{0.39}}$} & \best{41.66$_{\pm\text{0.15}}$} & \best{66.80$_{\pm\text{0.96}}$} \\
        % (a) & & & & & & $_{\pm\text{}}$ & $_{\pm\text{}}$ & $_{\pm\text{}}$ & $_{\pm\text{}}$ & $_{\pm\text{}}$ & $_{\pm\text{}}$ & $_{\pm\text{}}$ \\
        
        \bottomrule[0.15em]
    \end{tabular}{}
    }
    \caption{\textbf{This table shows the necessery of each part in Convolutional Fuser. }The model performs best when all parts are present. If remove any part of it, the model will perform worse. In this table, “\textsc{V2TEx}” means V2T Extractor, “\textsc{T2VEn}” means T2V Encoder, “\textsc{TrEn1}” means Transformer Encoder 1, “\textsc{ConBl}” means  Convolutional Blocks, “\textsc{TrEn2}” means Transformer Encoder 2. The best result in each column is highlighted in \best{bold}. }
    \label{tab:1dconvexperiment3}
\end{table*}
\endgroup

\subsubsection{Ablation Studies on Loop Decoder}

\begin{figure}[!h]
\centering
\includegraphics[width=0.3\columnwidth]{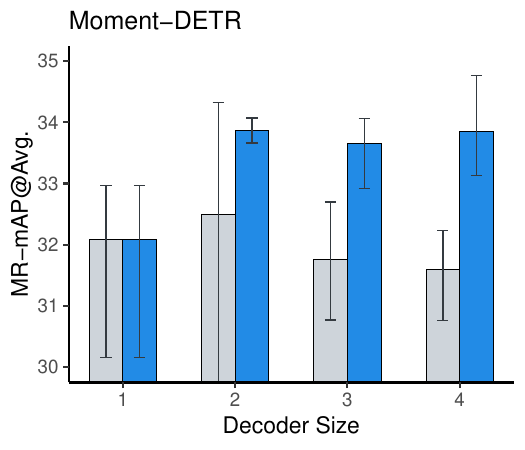}
\includegraphics[width=0.3\columnwidth]{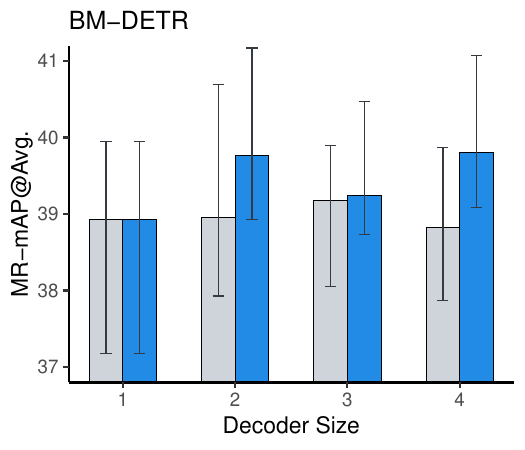}
\includegraphics[width=0.3\columnwidth]{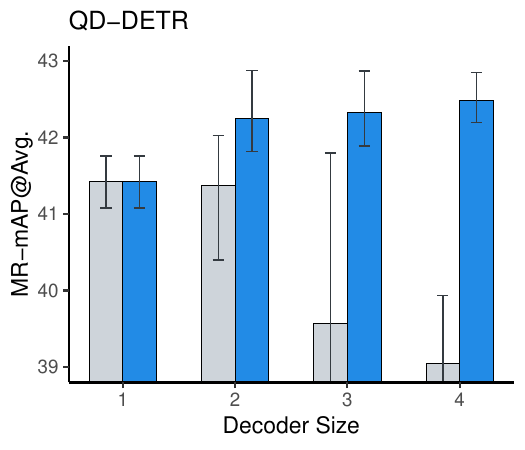}
\includegraphics[width=0.3\columnwidth]{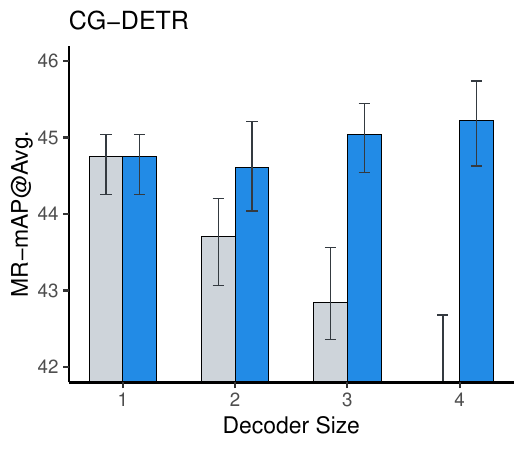}
\includegraphics[width=0.3\columnwidth]{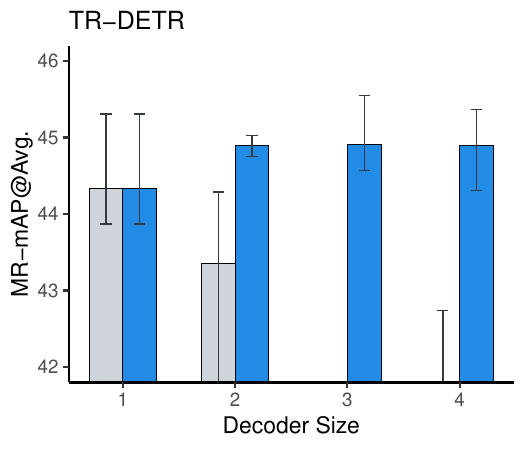}
\includegraphics[width=0.3\columnwidth]{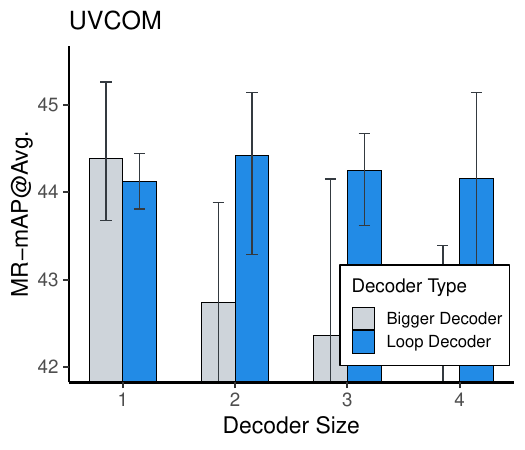}
\caption{\textbf{Loop Decoder as a plug-and-play method can help multiple models achieve better results without the risk of overfitting. }For Loop Decoder, the x-axis represents the number of Loop Decoder loops. For Bigger Decoder, the x-axis represents the multiple of the number of Decoder layers. This figure shows the same set of experiments as Table \ref{tab:loop_decoder_experiment1} 
% and Table \ref{tab:loop_decoder_experiment2}
. }
\label{fig:loop_decoder_experiment}
\end{figure}

Figure \ref{fig:loop_decoder_illustration} shows that Loop Decoder makes Video Moment Retrieval more accurate. It visualizes the Video Moment Retrieval results corresponding to the output of Loop Decoder at each loop. As the number of loops increases, the results are getting closer and closer to the ground truth. 
% Figure \ref{fig:loop_decoder_experiment}, Table \ref{tab:loop_decoder_experiment1} and Table \ref{tab:loop_decoder_experiment2} 
Figure \ref{fig:loop_decoder_experiment} and Table \ref{tab:loop_decoder_experiment1} 
shows the performance of Loop Decoder on multiple models. It demonstrates Loop Decoder as a plug-and-play method to improve the performance of multiple models. Through the Loop Decoder method, multimodal information is decoded more adequately. After using the Loop Decoder, the performance of the model is greatly improved. However, when using a bigger decoder of the same size, the model quickly overfits. 
We noticed that the effect of Loop Decoder on UVCOM~\cite{xiao2024bridging} is not as obvious as on other models. We speculate that this is because its Dual Branches Intra-Modality Aggregation affects the performance of our method. When we delete the Clip-Text Alignment method, the performance becomes better when the number of loops increases. But when we delete the Slot Attention method, the performance of the model becomes even better. 
% More details of the experiments are introduced in the Appendix. 

\begingroup
\fontsize{9}{11}\selectfont
\begin{table*}[!h]
    \centering
    \scalebox{0.47}{
    \setlength{\tabcolsep}{5.4mm}
        \begin{tabular}{ccccccccc}
        \toprule[0.15em]
        \multirow{3}{*}{\textbf{Decoder Type}} & \multirow{3}{*}{\textbf{Decoder Size}} & \multicolumn{5}{c}{\textbf{Video Moment Retrieval}} & \multicolumn{2}{c}{\textbf{Highlight Detection}} \\
        \cmidrule(l){3-7}  \cmidrule(l){8-9}
        & & \multicolumn{2}{c}{R1} & \multicolumn{3}{c}{mAP} & \multicolumn{2}{c}{$>=$Very Good} \\
        \cmidrule(l){3-4} \cmidrule(l){5-7}  \cmidrule(l){8-9}
        & & @0.5 & @0.7 & @0.5 & @0.75 & Avg. & mAP & HIT@1 \\
        
        \midrule[0.15em]
        \multicolumn{9}{l}{\textbf{Moment-DETR}~\cite{lei2021detecting} \source{NIPS'21}} \\
        \midrule

        \multirow{4}{*}{Bigger Decoder} & 1 & 53.63$_{\pm\text{1.59}}$ & 35.78$_{\pm\text{0.25}}$ & 54.99$_{\pm\text{1.33}}$ & 31.28$_{\pm\text{1.07}}$ & 32.09$_{\pm\text{1.01}}$ & \best{36.41$_{\pm\text{0.40}}$} & \best{56.92$_{\pm\text{0.98}}$} \\
        & 2 & 54.71$_{\pm\text{2.35}}$ & 35.84$_{\pm\text{2.97}}$ & 55.80$_{\pm\text{1.82}}$ & 31.39$_{\pm\text{2.35}}$ & 32.50$_{\pm\text{1.99}}$ & 36.17$_{\pm\text{0.63}}$ & 56.63$_{\pm\text{1.03}}$ \\
        & 3 & 53.65$_{\pm\text{0.73}}$ & 35.15$_{\pm\text{1.28}}$ & 55.11$_{\pm\text{0.68}}$ & 30.34$_{\pm\text{0.63}}$ & 31.75$_{\pm\text{0.73}}$ & 36.04$_{\pm\text{0.37}}$ & 56.53$_{\pm\text{0.98}}$ \\
        & 4 & 52.72$_{\pm\text{0.97}}$ & 35.14$_{\pm\text{0.86}}$ & 54.65$_{\pm\text{0.59}}$ & 30.64$_{\pm\text{0.59}}$ & 31.59$_{\pm\text{0.50}}$ & 35.78$_{\pm\text{0.29}}$ & 55.48$_{\pm\text{0.63}}$ \\

        \cmidrule(l){1-9}

        \multirow{4}{*}{Loop Decoder} & 1 & 53.63$_{\pm\text{1.59}}$ & 35.78$_{\pm\text{0.25}}$ & 54.99$_{\pm\text{1.33}}$ & 31.28$_{\pm\text{1.07}}$ & 32.09$_{\pm\text{1.01}}$ & \best{36.41$_{\pm\text{0.40}}$} & \best{56.92$_{\pm\text{0.98}}$} \\
        & 2 & 55.69$_{\pm\text{0.46}}$ & 37.63$_{\pm\text{0.55}}$ & \best{56.92$_{\pm\text{0.45}}$} & 32.93$_{\pm\text{0.36}}$ & \best{33.86$_{\pm\text{0.16}}$} & 36.37$_{\pm\text{0.41}}$ & 56.88$_{\pm\text{1.23}}$ \\
        & 3 & 54.93$_{\pm\text{1.14}}$ & 37.88$_{\pm\text{1.28}}$ & 56.19$_{\pm\text{0.76}}$ & 33.09$_{\pm\text{0.81}}$ & 33.66$_{\pm\text{0.45}}$ & 35.83$_{\pm\text{0.16}}$ & 55.30$_{\pm\text{0.60}}$ \\
        & 4 & \best{55.70$_{\pm\text{0.90}}$} & \best{38.24$_{\pm\text{0.85}}$} & 56.74$_{\pm\text{0.93}}$ & \best{33.19$_{\pm\text{0.70}}$} & 33.85$_{\pm\text{0.66}}$ & 36.10$_{\pm\text{0.43}}$ & 56.43$_{\pm\text{0.54}}$ \\

        \midrule
        \multicolumn{9}{l}{\textbf{BM-DETR}~\cite{jung2023overcoming} \source{arXiv'23}} \\
        \midrule

        \multirow{4}{*}{Bigger Decoder} & 1 & 60.90$_{\pm\text{0.79}}$ & 44.10$_{\pm\text{1.43}}$ & 60.91$_{\pm\text{0.96}}$ & 39.03$_{\pm\text{1.46}}$ & 38.93$_{\pm\text{0.98}}$ & - & - \\
        & 2 & 61.24$_{\pm\text{1.33}}$ & 44.13$_{\pm\text{1.42}}$ & 61.25$_{\pm\text{0.88}}$ & 39.21$_{\pm\text{1.06}}$ & 38.96$_{\pm\text{1.03}}$ & - & - \\
        & 3 & 60.72$_{\pm\text{0.54}}$ & 44.14$_{\pm\text{1.00}}$ & 61.14$_{\pm\text{0.43}}$ & 39.74$_{\pm\text{1.05}}$ & 39.18$_{\pm\text{0.68}}$ & - & - \\
        & 4 & 60.82$_{\pm\text{0.89}}$ & 43.73$_{\pm\text{0.78}}$ & 61.20$_{\pm\text{0.84}}$ & 38.97$_{\pm\text{0.80}}$ & 38.83$_{\pm\text{0.81}}$ & - & - \\

        \cmidrule(l){1-9}

        \multirow{4}{*}{Loop Decoder} & 1 & 60.90$_{\pm\text{0.79}}$ & 44.10$_{\pm\text{1.43}}$ & 60.91$_{\pm\text{0.96}}$ & 39.03$_{\pm\text{1.46}}$ & 38.93$_{\pm\text{0.98}}$ & - & - \\
        & 2 & 61.01$_{\pm\text{0.70}}$ & \best{44.80$_{\pm\text{0.96}}$} & 61.68$_{\pm\text{0.96}}$ & \best{40.34$_{\pm\text{0.93}}$} & 39.77$_{\pm\text{0.75}}$ & - & - \\
        & 3 & 61.42$_{\pm\text{1.20}}$ & 44.70$_{\pm\text{0.95}}$ & 61.71$_{\pm\text{1.06}}$ & 39.75$_{\pm\text{0.27}}$ & 39.24$_{\pm\text{0.63}}$ & - & - \\
        & 4 & \best{61.62$_{\pm\text{0.73}}$} & 44.76$_{\pm\text{1.48}}$ & \best{61.74$_{\pm\text{0.47}}$} & 40.14$_{\pm\text{0.92}}$ & \best{39.81$_{\pm\text{0.87}}$} & - & - \\

        \midrule
        \multicolumn{9}{l}{\textbf{QD-DETR}~\cite{moon2023query} \source{CVPR'23}} \\
        \midrule

        \multirow{4}{*}{Bigger Decoder} & 1 & 61.98$_{\pm\text{0.55}}$ & 47.30$_{\pm\text{0.69}}$ & 62.03$_{\pm\text{0.43}}$ & 41.96$_{\pm\text{0.66}}$ & 41.42$_{\pm\text{0.28}}$ & 38.92$_{\pm\text{0.30}}$ & 62.05$_{\pm\text{0.86}}$ \\
        & 2 & 62.59$_{\pm\text{0.70}}$ & 47.48$_{\pm\text{0.79}}$ & 62.02$_{\pm\text{0.75}}$ & 41.90$_{\pm\text{0.63}}$ & 41.37$_{\pm\text{0.54}}$ & 39.17$_{\pm\text{0.21}}$ & 62.18$_{\pm\text{0.67}}$ \\
        & 3 & 61.73$_{\pm\text{1.01}}$ & 45.38$_{\pm\text{1.63}}$ & 60.45$_{\pm\text{1.22}}$ & 39.99$_{\pm\text{1.75}}$ & 39.57$_{\pm\text{1.20}}$ & \best{39.24$_{\pm\text{0.23}}$} & \best{62.99$_{\pm\text{0.74}}$} \\
        & 4 & 60.96$_{\pm\text{0.43}}$ & 44.94$_{\pm\text{1.03}}$ & 59.92$_{\pm\text{0.40}}$ & 39.53$_{\pm\text{0.76}}$ & 39.04$_{\pm\text{0.77}}$ & 38.82$_{\pm\text{0.27}}$ & 61.39$_{\pm\text{0.93}}$ \\

        \cmidrule(l){1-9}

        \multirow{4}{*}{Loop Decoder} & 1 & 61.98$_{\pm\text{0.55}}$ & 47.30$_{\pm\text{0.69}}$ & 62.03$_{\pm\text{0.43}}$ & 41.96$_{\pm\text{0.66}}$ & 41.42$_{\pm\text{0.28}}$ & 38.92$_{\pm\text{0.30}}$ & 62.05$_{\pm\text{0.86}}$ \\
        & 2 & \best{62.80$_{\pm\text{0.80}}$} & 47.82$_{\pm\text{0.53}}$ & 62.97$_{\pm\text{0.49}}$ & 42.93$_{\pm\text{0.61}}$ & 42.26$_{\pm\text{0.37}}$ & 39.23$_{\pm\text{0.13}}$ & 62.83$_{\pm\text{0.85}}$ \\
        & 3 & 62.49$_{\pm\text{0.37}}$ & 47.57$_{\pm\text{0.44}}$ & \best{63.12$_{\pm\text{0.47}}$} & 42.68$_{\pm\text{0.36}}$ & 42.32$_{\pm\text{0.32}}$ & 38.94$_{\pm\text{0.19}}$ & 61.88$_{\pm\text{0.43}}$ \\
        & 4 & 62.45$_{\pm\text{0.86}}$ & \best{47.95$_{\pm\text{0.42}}$} & 63.03$_{\pm\text{0.41}}$ & \best{43.20$_{\pm\text{0.08}}$} & \best{42.48$_{\pm\text{0.24}}$} & 39.14$_{\pm\text{0.15}}$ & 62.67$_{\pm\text{0.54}}$ \\

        \midrule
        \multicolumn{9}{l}{\textbf{CG-DETR}~\cite{moon2023correlation} \source{arXiv'23}} \\
        \midrule

        \multirow{4}{*}{Bigger Decoder} & 1 & 65.92$_{\pm\text{0.22}}$ & 50.44$_{\pm\text{0.54}}$ & 65.50$_{\pm\text{0.22}}$ & 45.62$_{\pm\text{0.68}}$ & 44.76$_{\pm\text{0.26}}$ & 40.34$_{\pm\text{0.20}}$ & 65.12$_{\pm\text{0.64}}$ \\
        & 2 & 65.81$_{\pm\text{1.23}}$ & 50.09$_{\pm\text{0.54}}$ & 64.40$_{\pm\text{0.31}}$ & 44.17$_{\pm\text{0.51}}$ & 43.71$_{\pm\text{0.37}}$ & 40.33$_{\pm\text{0.25}}$ & 65.10$_{\pm\text{0.40}}$ \\
        & 3 & 65.30$_{\pm\text{0.87}}$ & 49.22$_{\pm\text{0.95}}$ & 63.50$_{\pm\text{0.85}}$ & 43.66$_{\pm\text{0.45}}$ & 42.85$_{\pm\text{0.45}}$ & \best{40.45$_{\pm\text{0.20}}$} & \best{65.61$_{\pm\text{0.50}}$} \\
        & 4 & 63.87$_{\pm\text{2.51}}$ & 46.92$_{\pm\text{4.35}}$ & 61.63$_{\pm\text{3.07}}$ & 41.30$_{\pm\text{3.54}}$ & 40.78$_{\pm\text{3.12}}$ & 40.37$_{\pm\text{0.11}}$ & 65.03$_{\pm\text{0.57}}$ \\

        \cmidrule(l){1-9}

        \multirow{4}{*}{Loop Decoder} & 1 & 65.92$_{\pm\text{0.22}}$ & 50.44$_{\pm\text{0.54}}$ & 65.50$_{\pm\text{0.22}}$ & 45.62$_{\pm\text{0.68}}$ & 44.76$_{\pm\text{0.26}}$ & 40.34$_{\pm\text{0.20}}$ & 65.12$_{\pm\text{0.64}}$ \\
        & 2 & 65.56$_{\pm\text{0.49}}$ & 50.58$_{\pm\text{0.59}}$ & 65.24$_{\pm\text{0.44}}$ & 45.29$_{\pm\text{0.59}}$ & 44.61$_{\pm\text{0.39}}$ & 40.10$_{\pm\text{0.28}}$ & 64.95$_{\pm\text{0.79}}$ \\
        & 3 & \best{66.26$_{\pm\text{0.64}}$} & \best{51.72$_{\pm\text{0.53}}$} & \best{65.74$_{\pm\text{0.39}}$} & 45.66$_{\pm\text{0.27}}$ & 45.05$_{\pm\text{0.32}}$ & 40.39$_{\pm\text{0.24}}$ & 65.53$_{\pm\text{0.47}}$ \\
        & 4 & 66.22$_{\pm\text{0.50}}$ & 51.04$_{\pm\text{0.36}}$ & 66.01$_{\pm\text{0.51}}$ & \best{45.95$_{\pm\text{0.52}}$} & \best{45.23$_{\pm\text{0.35}}$} & 40.33$_{\pm\text{0.13}}$ & 65.50$_{\pm\text{0.91}}$ \\

        \midrule
        \multicolumn{9}{l}{\textbf{TR-DETR}~\cite{sun2024tr} \source{AAAI'24}} \\
        \midrule

        \multirow{4}{*}{Bigger Decoder} & 1 & 66.56$_{\pm\text{1.06}}$ & 50.13$_{\pm\text{0.89}}$ & 65.70$_{\pm\text{0.79}}$ & 45.10$_{\pm\text{0.78}}$ & 44.33$_{\pm\text{0.51}}$ & 40.88$_{\pm\text{0.19}}$ & 65.54$_{\pm\text{0.45}}$ \\
        & 2 & 66.13$_{\pm\text{1.09}}$ & 49.54$_{\pm\text{1.04}}$ & 64.02$_{\pm\text{1.63}}$ & 43.86$_{\pm\text{1.11}}$ & 43.35$_{\pm\text{0.91}}$ & 40.61$_{\pm\text{0.37}}$ & 65.10$_{\pm\text{1.00}}$ \\
        & 3 & 64.18$_{\pm\text{1.22}}$ & 46.79$_{\pm\text{1.25}}$ & 62.46$_{\pm\text{0.96}}$ & 42.12$_{\pm\text{0.91}}$ & 41.33$_{\pm\text{0.85}}$ & \best{41.04$_{\pm\text{0.16}}$} & 65.28$_{\pm\text{0.57}}$ \\
        & 4 & 64.71$_{\pm\text{0.94}}$ & 46.92$_{\pm\text{1.35}}$ & 62.69$_{\pm\text{0.71}}$ & 42.26$_{\pm\text{1.25}}$ & 41.54$_{\pm\text{0.86}}$ & 40.85$_{\pm\text{0.10}}$ & \best{65.65$_{\pm\text{0.88}}$} \\

        \cmidrule(l){1-9}

        \multirow{4}{*}{Loop Decoder} & 1 & 66.56$_{\pm\text{1.06}}$ & 50.13$_{\pm\text{0.89}}$ & 65.70$_{\pm\text{0.79}}$ & 45.10$_{\pm\text{0.78}}$ & 44.33$_{\pm\text{0.51}}$ & 40.88$_{\pm\text{0.19}}$ & 65.54$_{\pm\text{0.45}}$ \\
        & 2 & 66.59$_{\pm\text{0.63}}$ & 51.10$_{\pm\text{0.40}}$ & 65.97$_{\pm\text{0.20}}$ & 45.42$_{\pm\text{0.20}}$ & 44.90$_{\pm\text{0.11}}$ & 40.93$_{\pm\text{0.29}}$ & 65.60$_{\pm\text{0.81}}$ \\
        & 3 & \best{67.12$_{\pm\text{0.36}}$} & 51.08$_{\pm\text{0.47}}$ & \best{66.35$_{\pm\text{0.46}}$} & \best{45.58$_{\pm\text{0.47}}$} & \best{44.92$_{\pm\text{0.33}}$} & 40.78$_{\pm\text{0.44}}$ & 65.37$_{\pm\text{0.80}}$ \\
        & 4 & 66.94$_{\pm\text{0.72}}$ & \best{51.40$_{\pm\text{0.51}}$} & 66.04$_{\pm\text{0.69}}$ & 45.22$_{\pm\text{0.64}}$ & 44.90$_{\pm\text{0.34}}$ & 40.85$_{\pm\text{0.11}}$ & 65.63$_{\pm\text{0.69}}$ \\

        % \midrule
        % \multicolumn{9}{l}{\textbf{Moment-DETR}~\cite{lei2021detecting}} \\
        % \midrule

        % \multirow{4}{*}{Bigger Decoder} & 1 & $_{\pm\text{}}$ & $_{\pm\text{}}$ & $_{\pm\text{}}$ & $_{\pm\text{}}$ & $_{\pm\text{}}$ & $_{\pm\text{}}$ & $_{\pm\text{}}$ \\
        % & 2 & $_{\pm\text{}}$ & $_{\pm\text{}}$ & $_{\pm\text{}}$ & $_{\pm\text{}}$ & $_{\pm\text{}}$ & $_{\pm\text{}}$ & $_{\pm\text{}}$ \\
        % & 3 & $_{\pm\text{}}$ & $_{\pm\text{}}$ & $_{\pm\text{}}$ & $_{\pm\text{}}$ & $_{\pm\text{}}$ & $_{\pm\text{}}$ & $_{\pm\text{}}$ \\
        % & 4 & $_{\pm\text{}}$ & $_{\pm\text{}}$ & $_{\pm\text{}}$ & $_{\pm\text{}}$ & $_{\pm\text{}}$ & $_{\pm\text{}}$ & $_{\pm\text{}}$ \\

        % \cmidrule(l){1-9}

        % \multirow{4}{*}{Loop Decoder} & 1 & $_{\pm\text{}}$ & $_{\pm\text{}}$ & $_{\pm\text{}}$ & $_{\pm\text{}}$ & $_{\pm\text{}}$ & $_{\pm\text{}}$ & $_{\pm\text{}}$ \\
        % & 2 & $_{\pm\text{}}$ & $_{\pm\text{}}$ & $_{\pm\text{}}$ & $_{\pm\text{}}$ & $_{\pm\text{}}$ & $_{\pm\text{}}$ & $_{\pm\text{}}$ \\
        % & 3 & $_{\pm\text{}}$ & $_{\pm\text{}}$ & $_{\pm\text{}}$ & $_{\pm\text{}}$ & $_{\pm\text{}}$ & $_{\pm\text{}}$ & $_{\pm\text{}}$ \\
        % & 4 & $_{\pm\text{}}$ & $_{\pm\text{}}$ & $_{\pm\text{}}$ & $_{\pm\text{}}$ & $_{\pm\text{}}$ & $_{\pm\text{}}$ & $_{\pm\text{}}$ \\

        \midrule
        \multicolumn{9}{l}{\textbf{UVCOM}~\cite{xiao2024bridging} \source{CVPR'24}} \\
        \midrule

        \multirow{4}{*}{Bigger Decoder} & 1 & 64.49$_{\pm\text{1.10}}$ & 49.94$_{\pm\text{0.72}}$ & 64.03$_{\pm\text{1.01}}$ & 44.88$_{\pm\text{0.53}}$ & 44.39$_{\pm\text{0.61}}$ & 39.94$_{\pm\text{0.18}}$ & 64.03$_{\pm\text{0.25}}$ \\
        & 2 & 64.37$_{\pm\text{0.33}}$ & 48.73$_{\pm\text{0.73}}$ & 62.85$_{\pm\text{0.88}}$ & 43.52$_{\pm\text{1.16}}$ & 42.73$_{\pm\text{0.81}}$ & 39.97$_{\pm\text{0.16}}$ & 63.42$_{\pm\text{1.09}}$ \\
        & 3 & 63.21$_{\pm\text{1.36}}$ & 47.52$_{\pm\text{1.81}}$ & 61.61$_{\pm\text{1.19}}$ & 43.23$_{\pm\text{1.35}}$ & 42.36$_{\pm\text{1.24}}$ & 39.81$_{\pm\text{0.13}}$ & 63.44$_{\pm\text{0.15}}$ \\
        & 4 & 62.68$_{\pm\text{2.90}}$ & 46.17$_{\pm\text{3.85}}$ & 60.74$_{\pm\text{2.57}}$ & 41.40$_{\pm\text{2.81}}$ & 40.99$_{\pm\text{2.84}}$ & 39.90$_{\pm\text{0.48}}$ & 63.72$_{\pm\text{0.64}}$ \\

        \cmidrule(l){1-9}

        \multirow{4}{*}{Loop Decoder} & 1 & 64.49$_{\pm\text{1.10}}$ & 49.94$_{\pm\text{0.72}}$ & 64.03$_{\pm\text{1.01}}$ & 44.88$_{\pm\text{0.53}}$ & 44.39$_{\pm\text{0.61}}$ & 39.94$_{\pm\text{0.18}}$ & 64.03$_{\pm\text{0.25}}$ \\
        & 2 & 63.81$_{\pm\text{0.37}}$ & 49.50$_{\pm\text{0.60}}$ & 63.60$_{\pm\text{0.38}}$ & 44.44$_{\pm\text{0.37}}$ & 44.03$_{\pm\text{0.24}}$ & 39.81$_{\pm\text{0.10}}$ & 63.37$_{\pm\text{0.38}}$ \\
        & 3 & 64.55$_{\pm\text{0.69}}$ & 50.04$_{\pm\text{0.95}}$ & 64.11$_{\pm\text{0.54}}$ & 45.41$_{\pm\text{1.90}}$ & 44.26$_{\pm\text{0.56}}$ & 39.94$_{\pm\text{0.17}}$ & 63.94$_{\pm\text{0.46}}$ \\
        & 4 & \best{64.65$_{\pm\text{0.61}}$} & \best{50.33$_{\pm\text{0.55}}$} & \best{64.17$_{\pm\text{0.46}}$} & 44.54$_{\pm\text{0.44}}$ & 44.18$_{\pm\text{0.29}}$ & 39.76$_{\pm\text{0.18}}$ & 63.69$_{\pm\text{0.80}}$ \\

        \cmidrule(l){1-9}

        \multirow{4}{*}{\makecell{Loop Decoder\\w/o Slot Attention}} & 1 & 64.36$_{\pm\text{0.44}}$ & 50.21$_{\pm\text{1.26}}$ & 63.99$_{\pm\text{0.27}}$ & \best{45.52$_{\pm\text{0.68}}$} & \best{44.72$_{\pm\text{0.53}}$} & 39.85$_{\pm\text{0.21}}$ & 63.82$_{\pm\text{1.05}}$ \\
        & 2 & 64.36$_{\pm\text{0.54}}$ & 50.06$_{\pm\text{0.80}}$ & 63.85$_{\pm\text{0.30}}$ & 45.15$_{\pm\text{0.68}}$ & 44.43$_{\pm\text{0.51}}$ & 39.88$_{\pm\text{0.24}}$ & \best{64.19$_{\pm\text{0.59}}$} \\
        & 3 & 64.00$_{\pm\text{0.64}}$ & 49.30$_{\pm\text{1.22}}$ & 63.73$_{\pm\text{0.54}}$ & 44.33$_{\pm\text{1.00}}$ & 43.93$_{\pm\text{0.63}}$ & 39.79$_{\pm\text{0.20}}$ & 63.54$_{\pm\text{0.74}}$ \\
        & 4 & 63.39$_{\pm\text{0.61}}$ & 49.02$_{\pm\text{0.50}}$ & 63.32$_{\pm\text{0.31}}$ & 43.88$_{\pm\text{0.52}}$ & 43.20$_{\pm\text{0.35}}$ & 39.35$_{\pm\text{0.18}}$ & 62.59$_{\pm\text{0.09}}$ \\

        \cmidrule(l){1-9}

        \multirow{4}{*}{\makecell{Loop Decoder\\w/o Clip-Text Alignment}} & 1 & 63.96$_{\pm\text{0.37}}$ & 49.73$_{\pm\text{0.42}}$ & 63.65$_{\pm\text{0.40}}$ & 44.55$_{\pm\text{0.47}}$ & 44.12$_{\pm\text{0.24}}$ & 39.82$_{\pm\text{0.20}}$ & 63.63$_{\pm\text{0.94}}$ \\
        & 2 & \best{64.65$_{\pm\text{0.53}}$} & 50.09$_{\pm\text{0.59}}$ & \best{64.17$_{\pm\text{0.72}}$} & 44.79$_{\pm\text{0.87}}$ & 44.42$_{\pm\text{0.61}}$ & 39.83$_{\pm\text{0.17}}$ & 63.90$_{\pm\text{0.41}}$ \\
        & 3 & 64.62$_{\pm\text{0.77}}$ & 50.10$_{\pm\text{0.40}}$ & 64.12$_{\pm\text{0.64}}$ & 44.74$_{\pm\text{0.44}}$ & 44.25$_{\pm\text{0.35}}$ & \best{39.98$_{\pm\text{0.15}}$} & 63.74$_{\pm\text{0.36}}$ \\
        & 4 & 63.85$_{\pm\text{0.78}}$ & 49.55$_{\pm\text{0.45}}$ & 63.60$_{\pm\text{1.13}}$ & 44.87$_{\pm\text{0.67}}$ & 44.15$_{\pm\text{0.82}}$ & 35.72$_{\pm\text{7.91}}$ & 63.54$_{\pm\text{0.67}}$ \\
        
        \bottomrule[0.15em]
    \end{tabular}{}
    }
    \caption{\textbf{Loop Decoder as a plug-and-play method can help multiple models achieve better results without the risk of overfitting like larger decoders. }The table categorizes the models by baselines. For Loop Decoder, Decoder Size represents the number of Loop Decoder loops, and for Bigger Decoder, Decoder Size represents the multiple of the number of Decoder layers. Slot Attention and Clip-Text Alignment are two methods used in the UVCOM~\cite{xiao2024bridging} model. The best result in each category of features in each column is highlighted in \best{bold}. This table shows the same set of experiments as Figure \ref{fig:loop_decoder_experiment}. }
    \label{tab:loop_decoder_experiment1}
\end{table*}
\endgroup

\clearpage

\section{Conclusion}

In this paper, we introduced a model LD-DETR for Video Moment Retrieval and Highlight Detection. We first introduced a plug-and-play method, \textit{Distill Align}, which mitigates the impact of overlapping semantic information. Then, we introduced \textit{Convolutional Fuser} which is more capable of capturing local information in multimodal features. Finally, we proposed a plug-and-play method, \textit{Loop Decoder}, which allows multimodal information to be more adequately decoded without causing overfitting. The superiority and effectiveness of our approach have been demonstrated on four public datasets. 

\clearpage

{\small
\bibliographystyle{plain}
\bibliography{references}
}

\end{document}